\documentclass[11pt,a4paper]{article}
\usepackage[hyperref]{emnlp2020}
\usepackage{times}
\usepackage{latexsym}
\usepackage{graphicx}
\usepackage{subcaption}

\usepackage{tabularx}
\newcolumntype{L}{>{\raggedright\arraybackslash}X}

\usepackage{microtype}

\aclfinalcopy 
\def\aclpaperid{8} 

\title{On the diminishing return of labeling clinical reports}

\author{Jean-Baptiste Lamare \\
  \texttt{jblamare@enlitic.com} \\\And
  Tobi Olatunji \\
  \texttt{tobi@enlitic.com} \\\And
  Li Yao \\
  \texttt{li@enlitic.com} \\}

\date{}

\begin{document}
\maketitle
\begin{abstract}
Ample evidence suggests that better machine learning models may be steadily obtained by training on increasingly larger datasets on natural language processing (NLP) problems from non-medical domains. Whether the same holds true for medical NLP has by far not been thoroughly investigated. This work shows that this is indeed not always the case. We reveal the somehow counter-intuitive observation that performant medical NLP models may be obtained with small amount of labeled data, quite the opposite to the common belief, most likely due to the domain specificity of the problem. We show quantitatively the effect of training data size on a fixed test set composed of two of the largest public chest x-ray radiology report datasets on the task of abnormality classification. The trained models not only make use of the training data efficiently, but also outperform the current state-of-the-art rule-based systems by a significant margin.
\end{abstract}

\section{Introduction}
It is commonly believed that neural network classifier performance increases as more data with labels are provided, if its capacity is properly optimized and regularized \citep{banko2001mitigating}. In natural language processing (NLP), pretraining on large TB-scale text corpora has become the standard practice in recent years with the emergence of BERT \citep{devlin2018bert} and GPT-2 \citep{radford2019language}, before being fine-tuned on task-specific targets. Furthermore, variants of top deep leanirng (DL) architectures have dominated several language task benchmarks due to their superior computational scalability and statistical capacity \citep{rajpurkar2017chexnet}. Modern neural networks typically employ a large number of parameters, offering large capacity and flexibility in capturing highly non-linear linguistic phenomena. As a model family with low bias and high variance, it naturally requires much more data to avoid the pitfall of overfitting.


The success of billion-parameter DL models trained on billion-word datasets have drastically transformed the landscape of NLP. Does the same trend hold true in the specific domain of medicine? After all, it is reasonable to expect a significant syntactic and semantic gap between everyday conversations (twitter feeds, news articles, blog posts) and formal medical vocabulary spoken or written in the context of clinical medicine. In fact, it typically requires decades of specialized medical training to excel in this highly demanding field. Such specialization is thus expected for any machine learning model that is trained to perform medical tasks, whether it is information retrieval \citep{goeuriot2016medical}, conversational agent \citep{laranjo2018conversational}, or disease extraction and classification \citep{chen2018deep}. 

This work investigates the impact of corpus size on the performance of state-of-the-art DL models on multi-label medical report classification tasks. We empirically demonstrate some of the unique properties of the medical language in clinical reports and how such domain-specific features lead to a surprisingly different scaling behavior as the training data size increases. Our main results on two public chest x-ray radiology report datasets consistently suggest that some of the classification tasks do not require copious amount of labeled data to achieve good performance, mostly due to the limited linguistic variation in its domain. Although per category analyses reveal slight variations, this phenomenon is consistently demonstrated across four DL model families presented. Performance between 6,000 and 30,000 reports remain counter-intuitively comparable, demonstrating diminishing returns in labeling effort. In addition, we show that, with a relatively small amount of data, DL models outperform both our private and state-of-the-art (SOTA) public ruled-based systems by a large margin.


\section{Related work}
\paragraph{Medical computer vision (CV)} Despite the obvious distinction between medical reports and medical images, the two modalities typically appear hand-in-hand in clinical environment. Diagnostic impression and recommendation of imaging studies are typically rendered in the format of structured and unstructured texts. For such reason, reports have become a popular and inexpensive way to derive large amount of labels for machine learning CV tasks at scale \citep{attaluri2018efficient, olatunji2019caveats,olatunji2019learning}. In fact, labels obtained this way have been widely used to form large training set for fundamental CV tasks such as triage, detection and segmentation in \citet{yao2017learning,yao2018weakly, yao2019strong}. Therefore, accurate  medical NLP brings significant benefit to the development of CV models as a whole.
\paragraph{Non-medical NLP} The early work of \citet{banko-brill-2001-scaling} demonstrates a linear performance gain of an NLP disambiguation task when doubling the amount of training corpus on simple machine learning (ML) linear classifiers. Recent NLP advance pushes the envelop much further by leveraging web-scale data -- for instance, the Common Crawl project \footnote{\url{http://commoncrawl.org}} that produces 20TB of textual data from the Internet each month. To cope with such a scale, large models with billions of parameters based on the variants of BERT \citep{devlin2018bert}, MT-DNN \citep{liu2019multi}, GPT-2 \citep{radford2019language}, XL-Net \citep{yang2019xlnet} have emerged with sometimes near-human performance on selected language tasks. It is commonly recognized that better performance can be achieved by training larger models on larger datasets.

\paragraph{Non-ML medical NLP} Although traditional NLP performance has improved on medical tasks over time, it doesn't lend itself to the scale of significant performance improvements seen with models trained on datasets several terabytes in size \citep{biobert}. Results shown in a 1999 paper \citep{taira1999statistical} on statistical NLP for medical reports, a 2006 paper \citep{meystre2006natural} extracting predefined problems from clinical notes, a 2013 paper mining FDA drug labels \citep{li2013mining}, and a 2019 dataset using Chexpert labeler \citep{irvin2019chexpert} demonstrate gradual performance improvements that still reflect the high precision, low recall phenomenon observed with non-ML medical NLP tools.  

\paragraph{ML medical NLP} Nonetheless, machine learning based NLP models have been explored where a reasonable amount of electronic medical records (EMR) are obtained privately. \citet{chen2018deep} uses a convolutional neural network (CNN) trained on 2500 thoracic computed tomography (CT) reports to identify pulmonary embolism findings with high AUCs. \citet{lee2019automatic} uses a corpus of 3032 musculoskeletal x-ray reports to train a recurrent neural network (RNN) to identify fracture and non-fracture cases with high precision and recall. The work from \citet{rajkomar2018scalable} represents one of the largest studies on EMR where free-text notes from doctors, nurses and other providers from 216K patients are used to predict mortality, readmission, and length of hospital stay. \citet{liu2019clinically} uses hundreds of thousands of chest x-ray reports, but focuses on the task of report generation instead of classification. 

\paragraph{Corpus Size}
\citet{roberts2016assessing} evaluates the impact of combining 6 clinical and non-clinical corpora on similarity of word embeddings in the clinical domain. Results showed task-dependent performance variations. A study from \citet{ahmed2018resource} on NER in a low resource language showed improved performance when 3 datasets were combined. In line with the dominant trend, \citet{banko2001mitigating} applied machine learning
techniques to the task of confusion set disambiguation, using three
orders of magnitude more training data than previously been applied to the problem. They significantly reduced the error rate simply by adding more training data. Even with a billion words, the learners continued to benefit from additional training data. In contrast, \citet{curran2002very} confronts this claim showing that although convergence behaviour on unigram probability estimates improves when using up to one billion words, for some words, no such convergence occurs.

In contrast to the aforementioned studies, this work focuses on analyzing the comparative performance of NLP classifiers on radiology reports in a multi-class setting with respect to different amounts of training data.

\section{Experiments}
The primary goal of this study is to empirically examine the impact of increasing and decreasing the size of the training data to the quality of DL models produced, and to compare their performance with state-of-the-art rule-based methods. Thus in all following experiments, we freeze the test set while using different sizes of training and validation set to tune the DL models. In particular, we establish incrementally bigger training data by randomly sampling a larger percentage from the entire training set. As a baseline for performance comparison, we also introduce two rule-based classifiers whose performance is independent of the training data sizes.
\subsection{Dataset}

\paragraph{Data} Our data consists of several datasets coming from different international sources, both public and private. On the public side, we use a subset of the MIMIC-CXR dataset \cite{johnson2019mimic}, a large dataset of 377,110 chest x-rays associated with 227,827 imaging studies and corresponding radiology reports. The reports are also provided with labels obtained with the CheXpert labeling tool \cite{irvin2019chexpert}. 
We also incorporate the OpenI dataset \cite{demner2012design}, a collection of 7,470 chest x-rays with 3,955 radiology reports. On the private side, we add about 21,000 reports from our in-house chest x-ray datasets as additional training set.

\paragraph{Splits} We only use public datasets to create the test set. First we take the original MIMIC-CXR test split. In addition, we randomly sample half of the OpenI dataset to be used in the test split. We then use the remaining 33,000 reports for training and validation, randomly sampling 10\% of it for validation and leaving about 30,000 reports for training. The number of reports per split and provenance are available in table \ref{tab_split}.

\begin{table*}
\centering

\begin{tabular}{|l||c|c|c||c|}
\hline
Source           & Training   & Validation  & Testing  & Total per source \\ \hline
MIMIC-CXR        & 9,349      & 1,039       & 3,088    & 13,476           \\ \hline
OpenI            & 1,676      & 186         & 1,862    & 3,724            \\ \hline
Private data     & 18,900     & 2,100       & 0        & 21,000           \\ \hline \hline
Total per split  & 29,925     & 3,325       & 4,950    & 38,200           \\ \hline

\end{tabular}
\caption{Number of reports per source and split in the full dataset. The ablation sampling was then done with random sampling out of the training/validation set while the test set is frozen.}
\label{tab_split}
\end{table*}

\paragraph{Labels} A team of experts manually provided labels from scratch on the full train/validation/test dataset. Each report was labeled by a single expert, following a private labeling scheme. 
We also modified the Chexpert labeling scheme to align with label definitions and labeling guidelines used by our team of experts. We merged \textit{Consolidation} and \textit{Pneumonia} reflecting the significant visual overlap between both labels, we excluded \textit{Edema}, \textit{Cardiomegaly} and \textit{No Finding}. These modifications make our and Chexpert's labeling schemes comparable. We show in Table \ref{tab_abn} the label counts for all of the abnormalities in the training and test sets.

\begin{table*}
\centering

\begin{tabular}{|l||c|c||c|c||c|c||}
\hline
                        & \multicolumn{2}{c||}{Training} & \multicolumn{2}{c||}{Test-MIMIC} & \multicolumn{2}{c||}{Test-OpenI} \\ 
                        & +            & -           & +            & -         & +     & -       \\ \hline
atelectasis             & 6,804        & 23,121      & 1,062        & 2,026     & 173   & 1,689   \\ \hline
consolidation/pneumonia & 7,178        & 22,747      & 721          & 2,367     & 115   & 1,747   \\ \hline
enlarged cardiomediastinum & 5,100     & 24,825      & 1,196        & 1,892     & 215   & 1,647   \\ \hline
fracture                & 1,857        & 28,068      & 182          & 2,906     & 53    & 1,809   \\ \hline
lung lesion             & 2,875        & 27,050      & 215          & 2,873     & 252   & 1,610   \\ \hline
lung opacity            & 10,581       & 19,344      & 851          & 2,237     & 730   & 1,132   \\ \hline
pleural effusion        & 6,582        & 23,343      & 1,293        & 1,795     & 90    & 1,772   \\ \hline
pneumothorax           & 1,057        & 28,868      & 90           & 2,998     & 12    & 1,850   \\ \hline
pleural other           & 1,987        & 27,938      & 194          & 2,894     & 34    & 1,828   \\ \hline
support devices         & 9,789        & 20,136      & 1,278        & 1,810     & 169   & 1,693   \\ \hline
\end{tabular}
\caption{Label counts for all 10 of the selected abnormalities}
\label{tab_abn}
\end{table*}

\subsection{Training}
\paragraph{DL models}
We train four types of NLP multilabel classifiers, including three relatively small architectures without pretraining and a bigger pretrained BERT model:
\begin{itemize}
    \item Bidirectional LSTM with attention and dropout. We use word embeddings of size 128 and 128 units. 
    \item CNN \cite{kim2014cnn} following the cited architecture with embedding size 128, 128 filters, and kernel sizes 3, 4 and 5.
    \item RCNN \cite{rcnn2015} also with embeddings of size 128 and 128 units.
    \item BERT \cite{devlin2018bert}, with the pretrained weights from BioBERT \cite{biobert}
\end{itemize}
All models were trained with the Adam optimizer for 20 epochs, with early stopping and a learning rate of 1e-3 except for BERT which had a 5e-5 learning rate (default values). Before training, reports were cleaned by removing headers and footers using keywords. For non-BERT models, reports were preprocessed with the NLTK tokenizer \footnote{https://www.nltk.org/}, while for BERT we used the original WordPiece tokenizer \cite{wordpiece}.

\paragraph{Non-DL rule-based medical NLP baselines}

Domain-specific heavily-engineered hand-crafted rule-based NLP tools yield binary outputs for the presence (1) or absence (0) of abnormalities \cite{olatunji2019caveats} \cite{hassanpour2016information} \cite{attaluri2018efficient}. SOTA tools typically include additional capabilities that express the associated degree of uncertainty in the report. \cite{peng2018negbio}, \cite{rajpurkar2017chexnet} \cite{savova2010mayo}.  
This is primarily driven by the high cost of collecting expert-level human annotations. Bootstrapping labels from radiology reports has therefore gained significant attention in recent years. As illustrated in  \cite{olatunji2019caveats}, these hand-crafted tools generally suffer from low recall, a problem we promptly address. Our domain-specific rule-based NLP tool automatically extracts labels from reports in 3 main steps -- extraction, classification and aggregation-- similar to Chexpert \cite{irvin2019chexpert}. Classification (Negation and Uncertainty detection) rules were also designed on the universal dependency parse of the report. We, however pursue alternate strategies for mention extraction, negation, uncertainty detection and aggregation that yield high recall and comparable precision, like \citet{olatunji2019learning}.

\paragraph{Evaluation metric}

Unlike rule-based non-DL methods that predict directly a discrete output for each abnormality (e.g., 1 for existence and 0 for absence), DL models output a continuous score indicating the probability of its presence. Typically one needs to decide on binarization threshold to convert DL outputs to discrete decisions which requires additional domain or application specific prior, especially in the field of medical diagnosis where true positives and false negatives are associated with different risk factors. Without such information a priori, one could still use threshold-independent metrics to evaluate model performance. Therefore, for this work, we choose precision recall curve (PRC) and area under precision recall curve (AUC-PR). Recall is equivalent to the standard sensitivity metric in medicine. Precision is more sensitive to false positives than the commonly used specificity when there are large amount of easy-to-classify negatives.

\section{Results}
\subsection{Global Results Analysis}
\begin{figure*}
\centering
\begin{subfigure}{.25\textwidth}
\centering
\includegraphics[width=\linewidth]{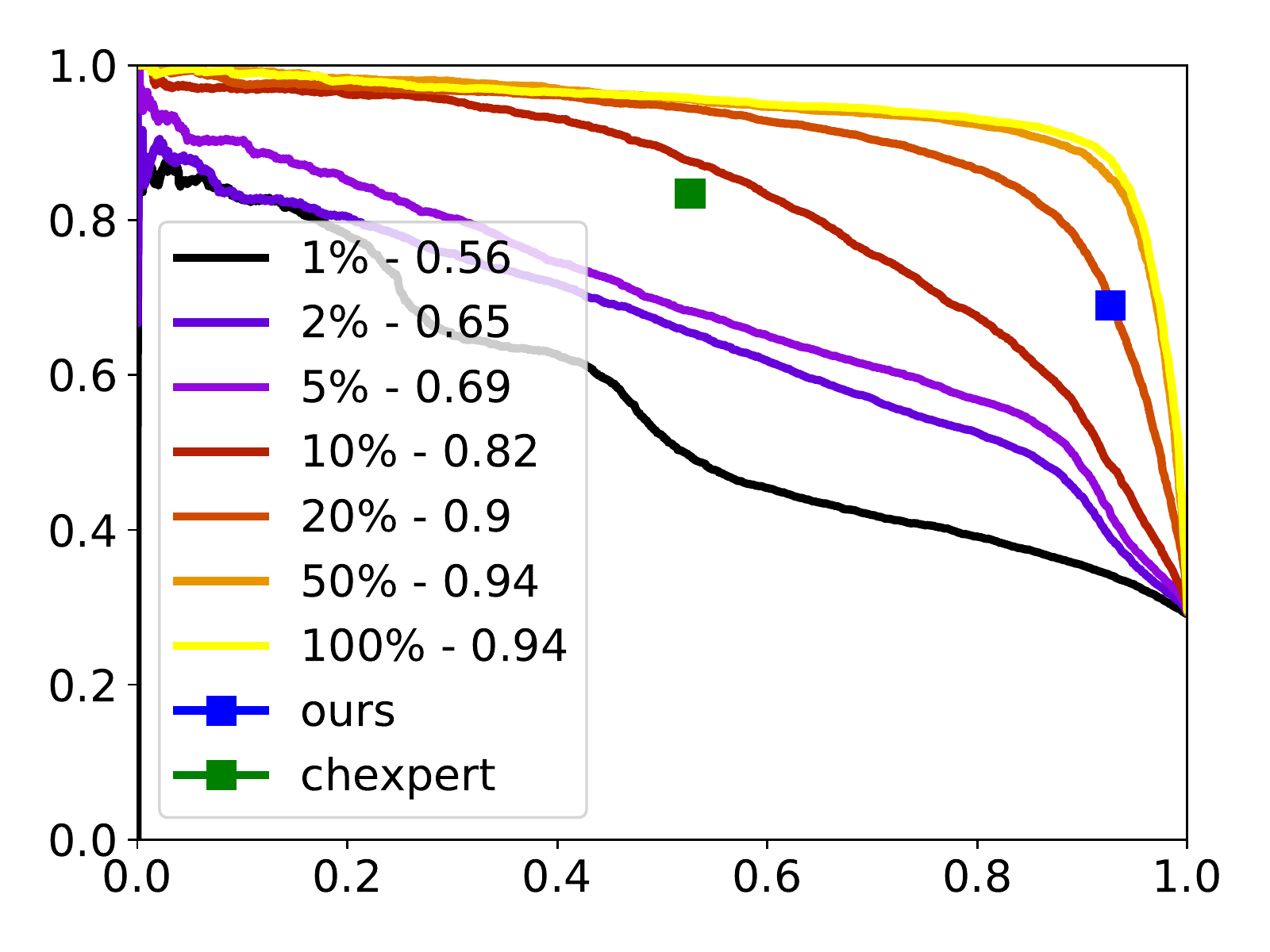}
\caption{LSTM on MIMIC}
\label{sfig:lstm_mimic}
\end{subfigure}%
\begin{subfigure}{.25\textwidth}
\centering
\includegraphics[width=\linewidth]{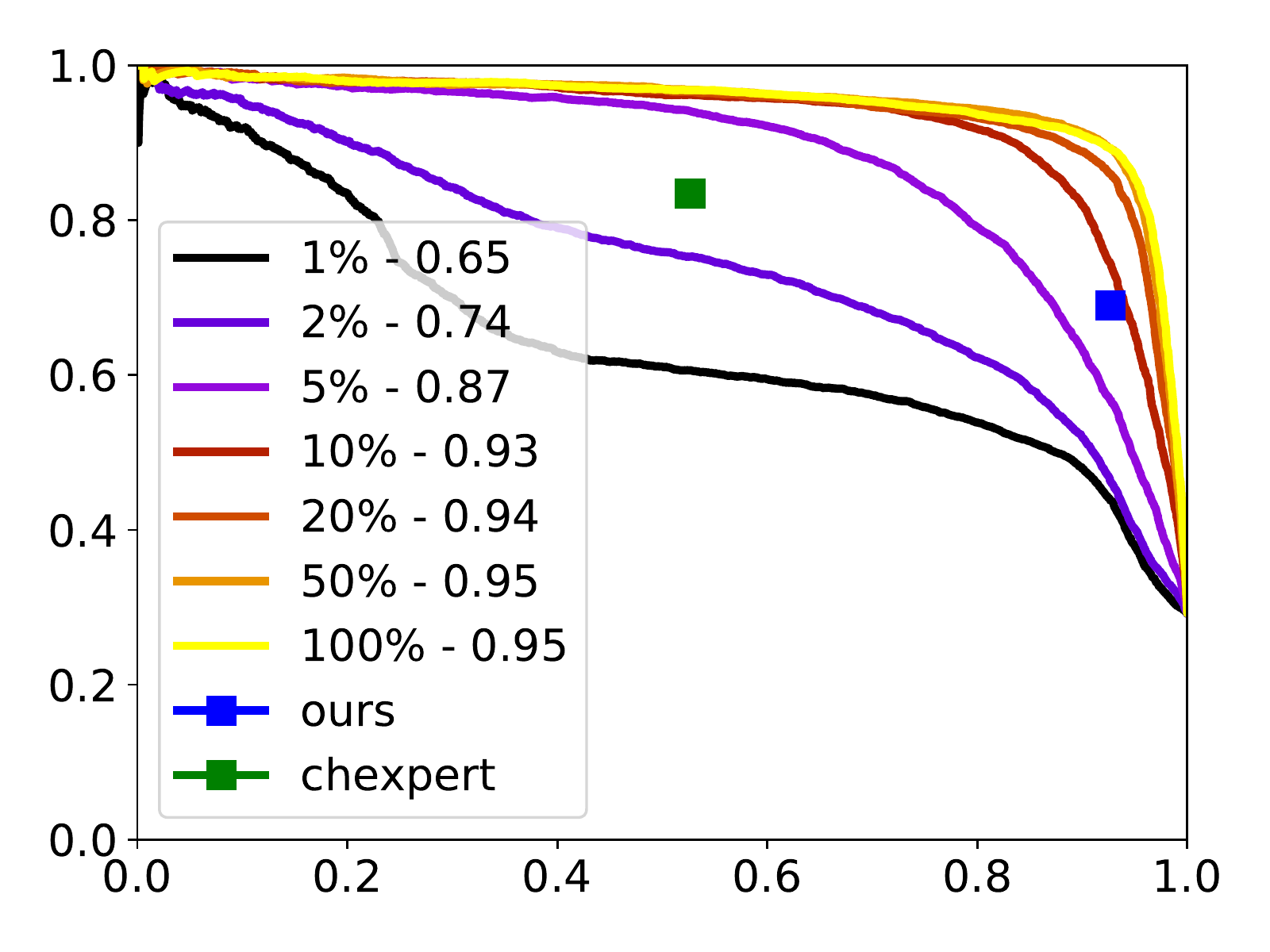}
\caption{CNN on MIMIC}
\label{sfig:cnn_mimic}
\end{subfigure}%
\begin{subfigure}{.25\linewidth}
\centering
\includegraphics[width=\linewidth]{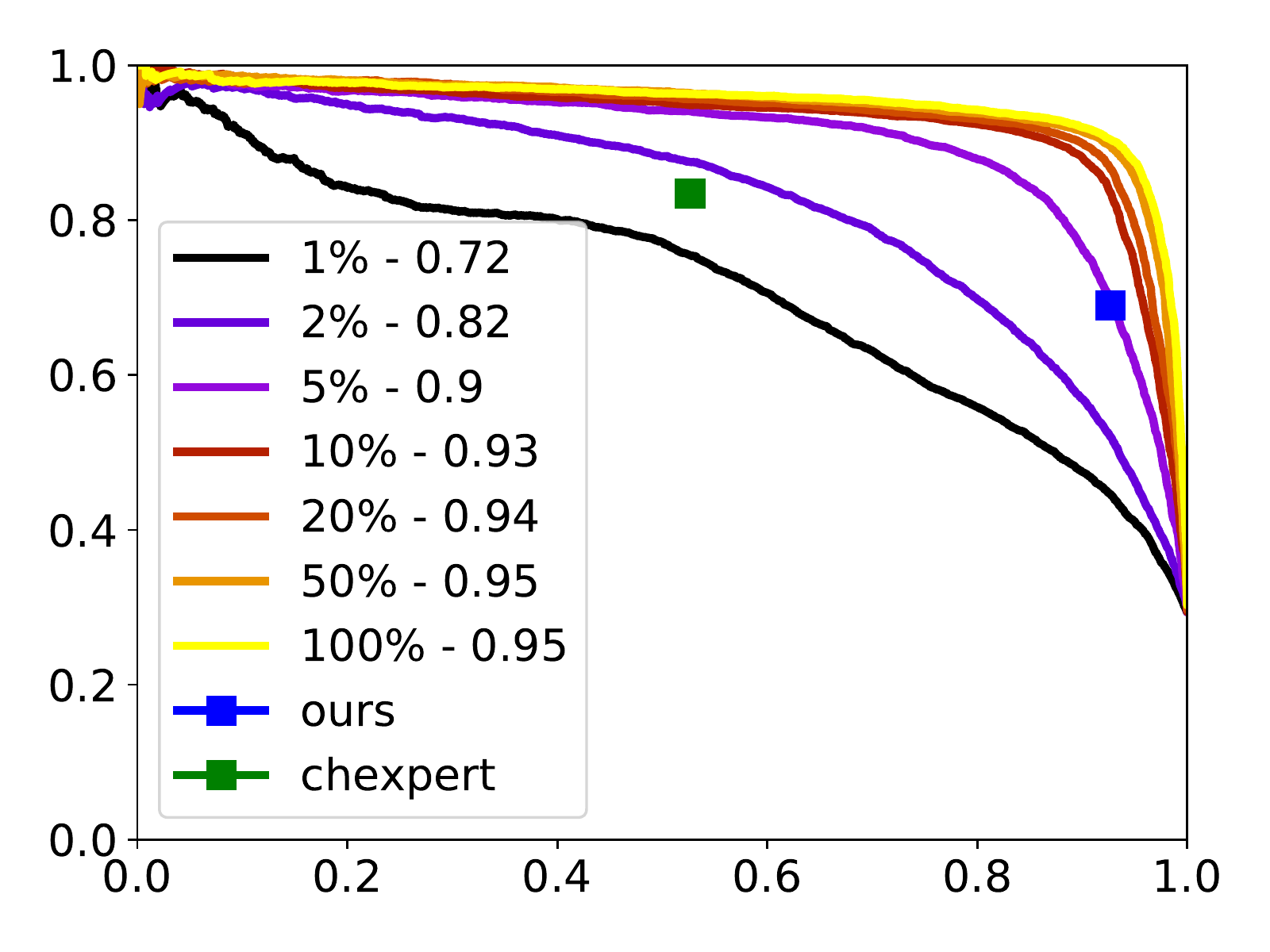}
\caption{RCNN on MIMIC}
\label{sfig:rcnn_mimic}
\end{subfigure}%
\begin{subfigure}{.25\textwidth}
\centering
\includegraphics[width=\linewidth]{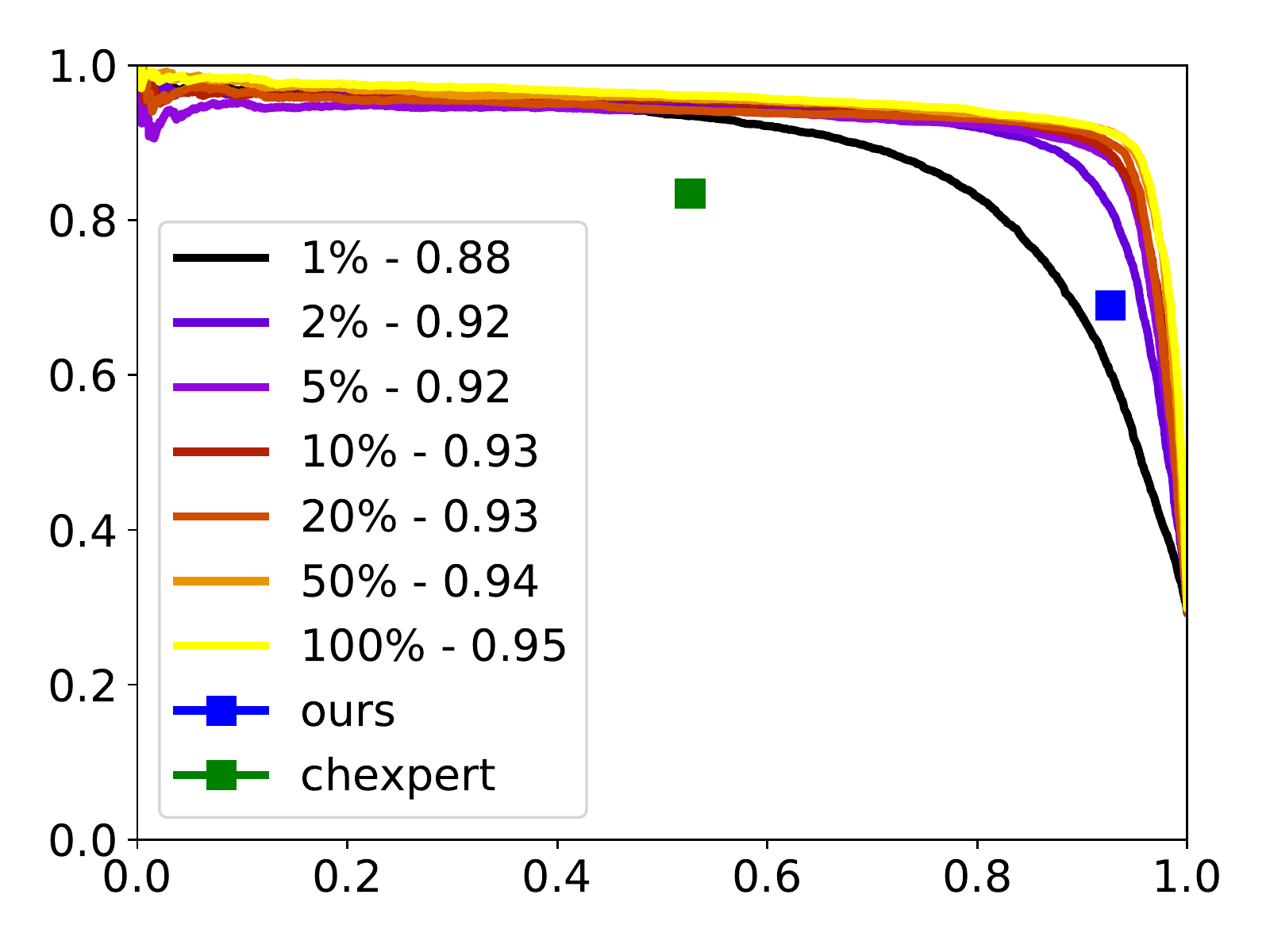}
\caption{BERT on MIMIC}
\label{sfig:bert_mimic}
\end{subfigure}\par\medskip
\begin{subfigure}{.25\textwidth}
\centering
\includegraphics[width=\linewidth]{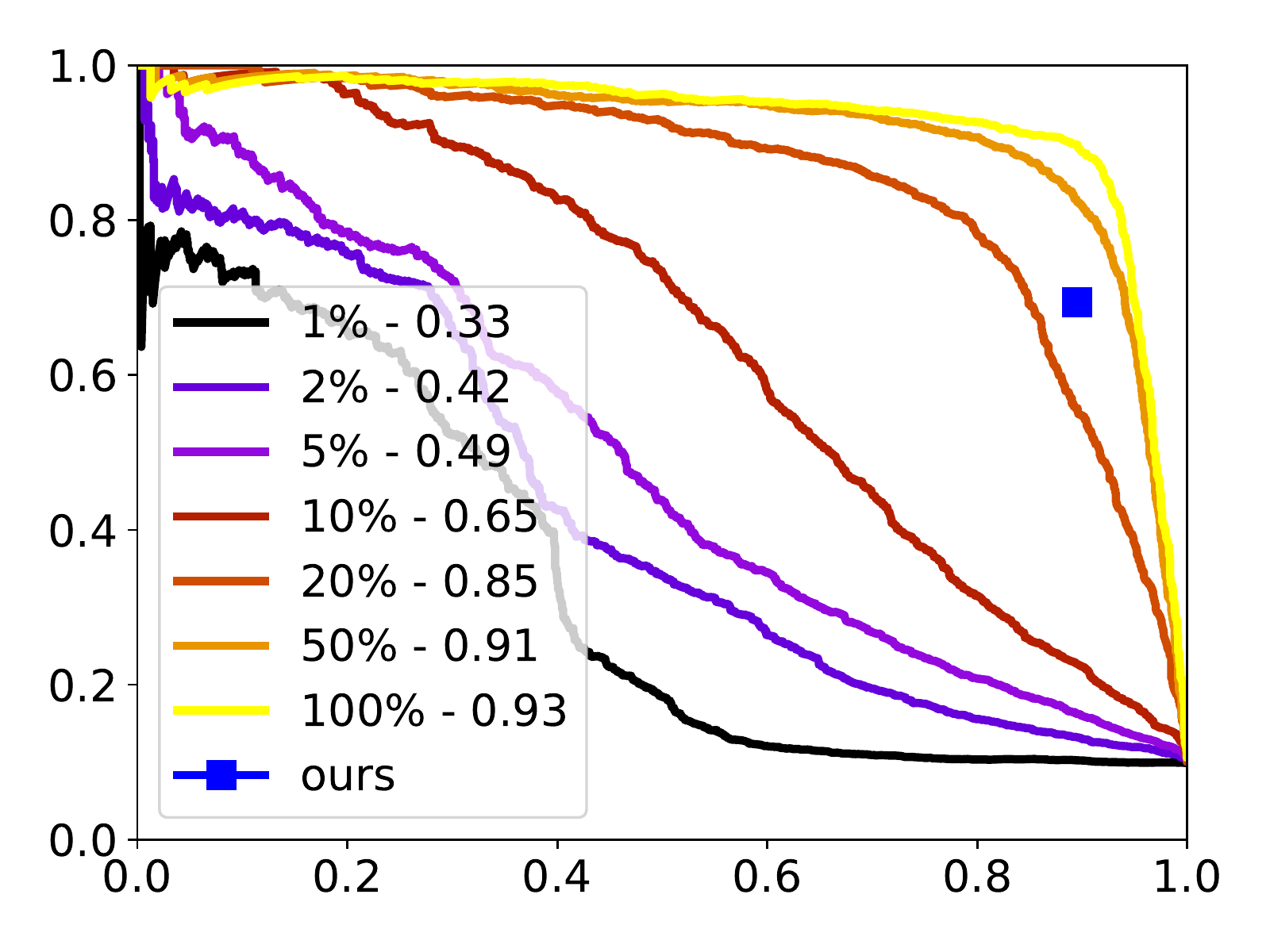}
\caption{LSTM on OpenI}
\label{sfig:lstm_openi}
\end{subfigure}%
\begin{subfigure}{.25\linewidth}
\centering
\includegraphics[width=\linewidth]{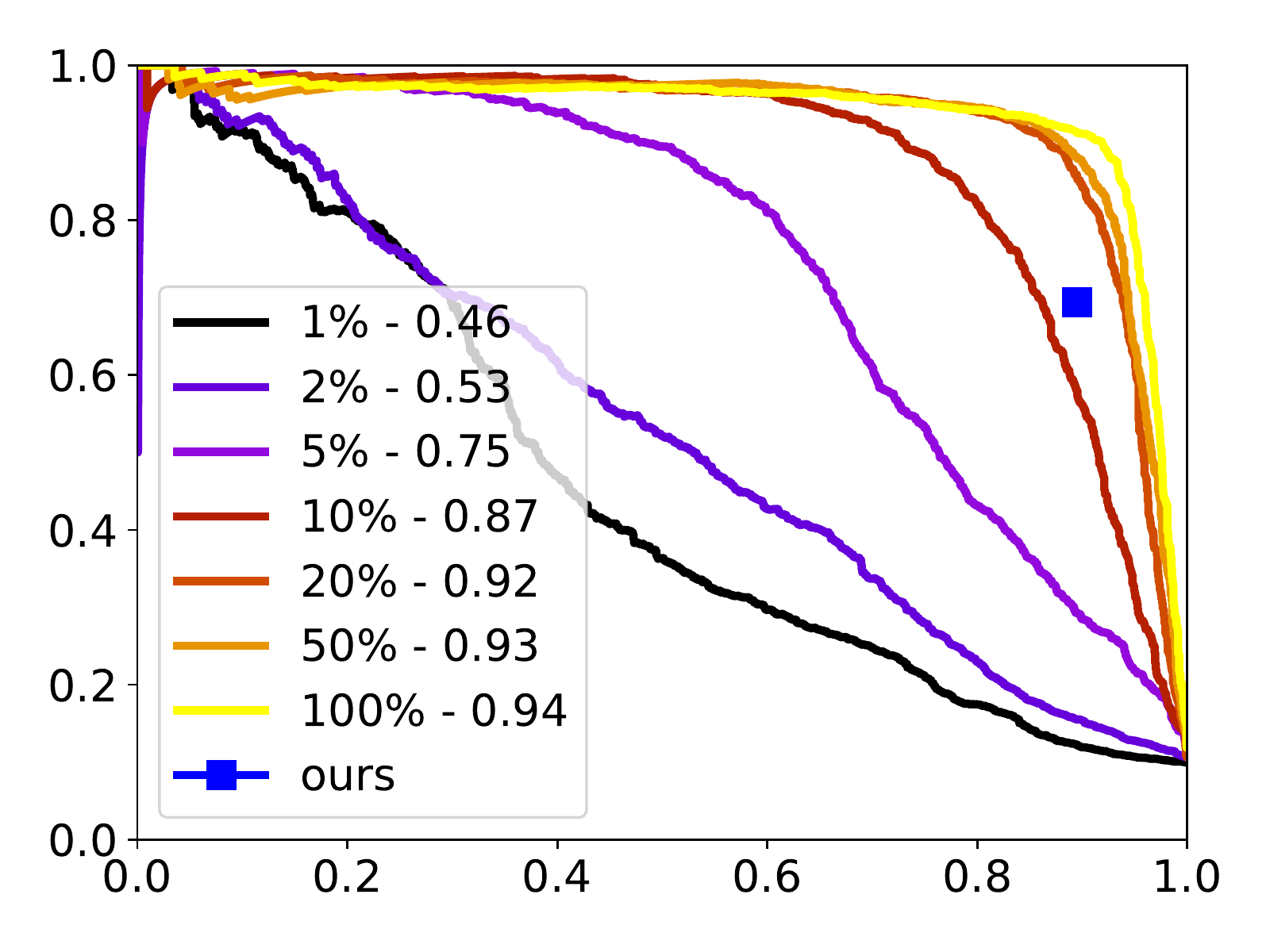}
\caption{CNN on OpenI}
\label{sfig:cnn_openi}
\end{subfigure}%
\begin{subfigure}{.25\linewidth}
\centering
\includegraphics[width=\linewidth]{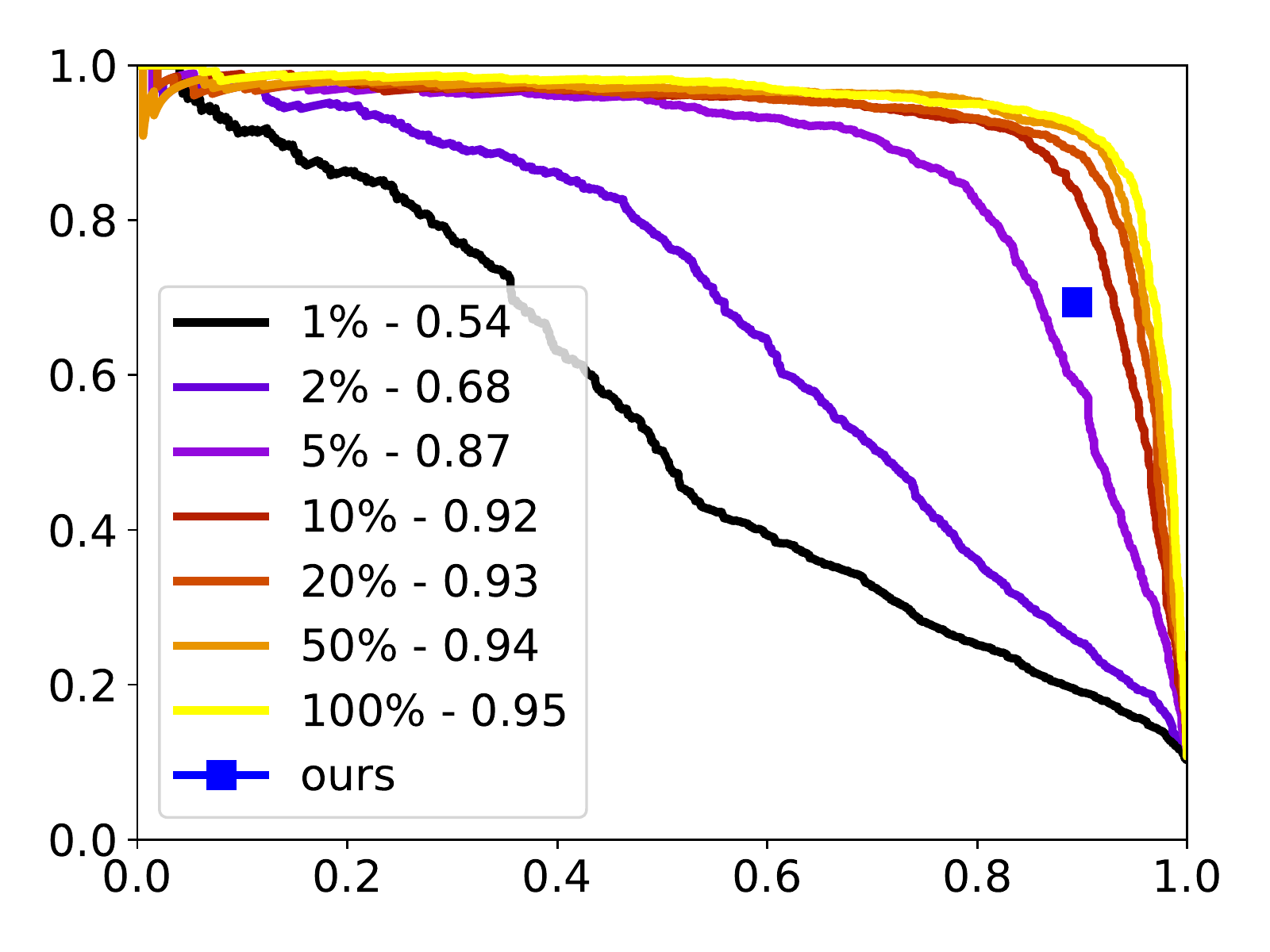}
\caption{RCNN on OpenI}
\label{sfig:rcnn_openi}
\end{subfigure}%
\begin{subfigure}{.25\textwidth}
\centering
\includegraphics[width=\linewidth]{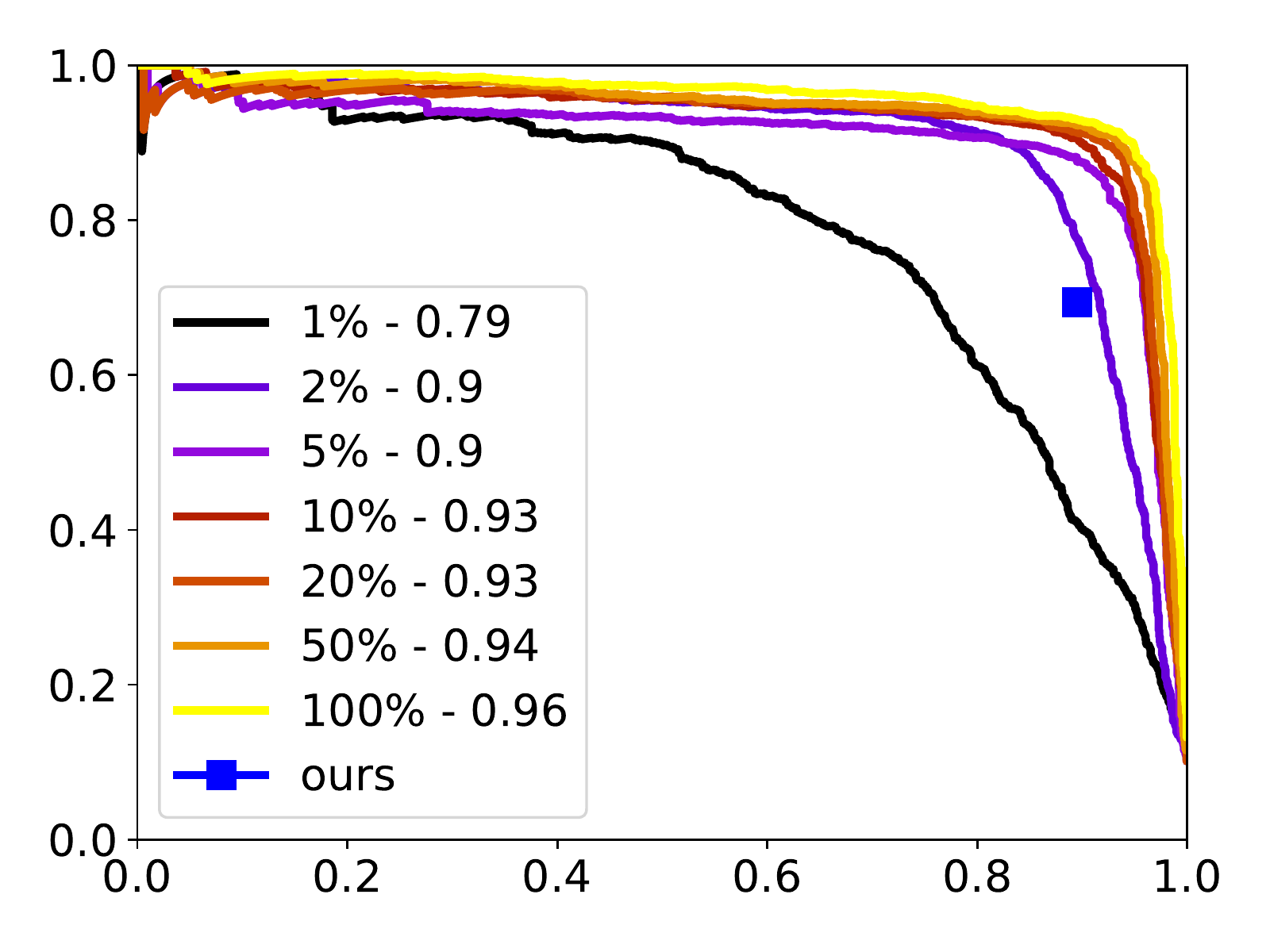}
\caption{BERT on OpenI}
\label{sfig:bert_openi}
\end{subfigure}\par\medskip

\caption{PR curves on MIMIC (first row) and OpenI (second row) test sets with different data sizes and comparison with rule-based performance. As commonly shown, the y axis is Precision while the x axis is Recall.}
\label{fig:micro}
\end{figure*}

A quantitative summary using the AUC-PR can be found in table \ref{tab_results} for all four DL models. We also show some of the Precision-Recall curves as well as the rule-based baselines in Figure \ref{fig:micro}. Contrary to the general trend of performance increase with dataset size, the results demonstrate an interesting phenomenon. Despite the fact that the full dataset size is 30,000, by no means comparable to datasets with billions of tokens, 4 different model architectures (CNN, RNN, RCNN, BioBERT) achieve over 0.94 overall AUC-PR on the MIMIC test set, outperforming SOTA held by domain-specific hand-crafted rule-based systems. As seen in Table \ref{tab_results}, performance remains comparable as dataset size drops from 100\% to about 20\% where AUC-PRs drop below 0.90, at which point DL algorithms start to approach the performance of domain specific rule-based tools. 

This is counter-intuitive. Without rigorously investigating the effect of dataset size, dominant rule-based systems and ML benchmarks seem to suggest medical NLP is a more difficult task. Our findings suggest otherwise. Additionally, a 2001 study \citep{campbell2001comparing} comparing syntactic complexity in medical and non-medical corpora shows that the syntax of medical language shows less variation than non-medical language and is likely simpler. 

Our experiments demonstrate that with only about 6000 reports (20\% of the data), multiple DL model architectures achieve and sustain micro AUC-PR over 0.90 across multiple label categories while surpasses the non-DL SOTA baselines.

On the OpenI test set, results are slightly worse across the board for DL but still conform to the general trend. It is important to note that performance of the rule based method remains fairly consistent despite the obvious drop in DL performance. This highlights one of the uncelebrated strengths of non-ML medical NLP.

\begin{table*}
\centering
\begin{tabular}{|l||c|c|c|c||c|c|c|c||}
\hline
Training data        & \multicolumn{4}{c||}{Test-MIMIC} & \multicolumn{4}{c||}{Test-OpenI} \\ 
Percentage      & RNN       & CNN       & RCNN       & BERT      & RNN       & CNN       & RCNN       & BERT       \\ \hline
1\%             & 0.562     & 0.650     & 0.717      & 0.878     & 0.332     & 0.461     & 0.543      & 0.788      \\ \hline
2\%             & 0.653     & 0.745     & 0.821      & 0.920     & 0.423     & 0.530     & 0.677      & 0.900      \\ \hline
5\%             & 0.693     & 0.874     & 0.902      & 0.918     & 0.492     & 0.749     & 0.865      & 0.905      \\ \hline
10\%            & 0.819     & 0.925     & 0.928      & 0.933     & 0.653     & 0.875     & 0.916      & 0.928      \\ \hline
20\%            & 0.902     & 0.940     & 0.937      & 0.928     & 0.852     & 0.924     & 0.926      & 0.929      \\ \hline
50\%            & 0.938     & 0.948     & 0.946      & 0.944     & 0.914     & 0.928     & 0.938      & 0.942      \\ \hline
100\%           & 0.941     & 0.948     & 0.950      & 0.949     & 0.927     & 0.937     & 0.950      & 0.956      \\ \hline
\end{tabular}
\caption{AUC-PR evolution on the test sets with different dataset sizes}
\label{tab_results}
\end{table*}

\subsection{Per Category Analysis}

\begin{figure*}
\centering
\begin{subfigure}{.25\textwidth}
\centering
\includegraphics[width=\linewidth]{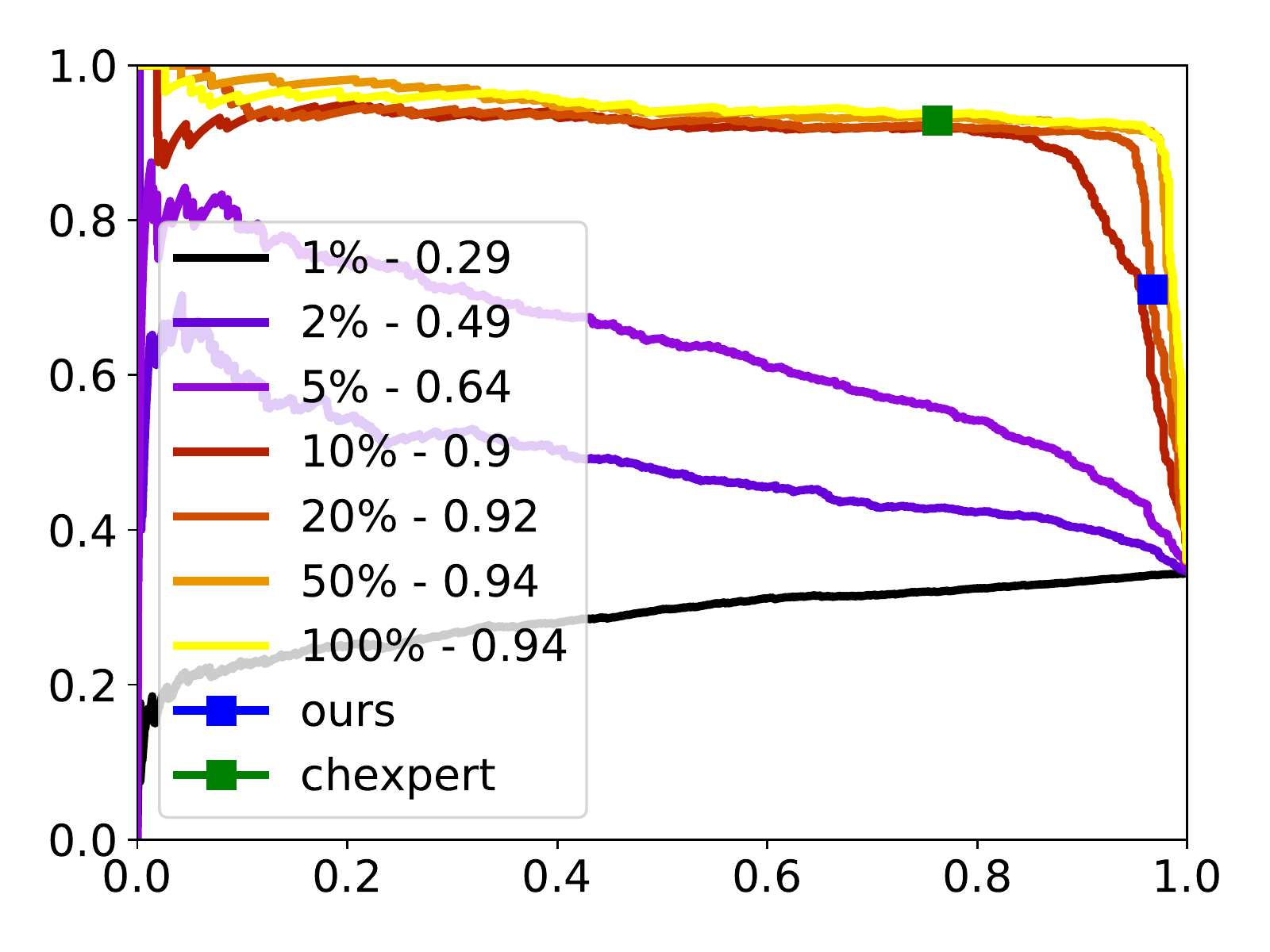}
\caption{Atelectasis}
\label{sfig:rnn_atelectasis}
\end{subfigure}%
\begin{subfigure}{.25\textwidth}
\centering
\includegraphics[width=\linewidth]{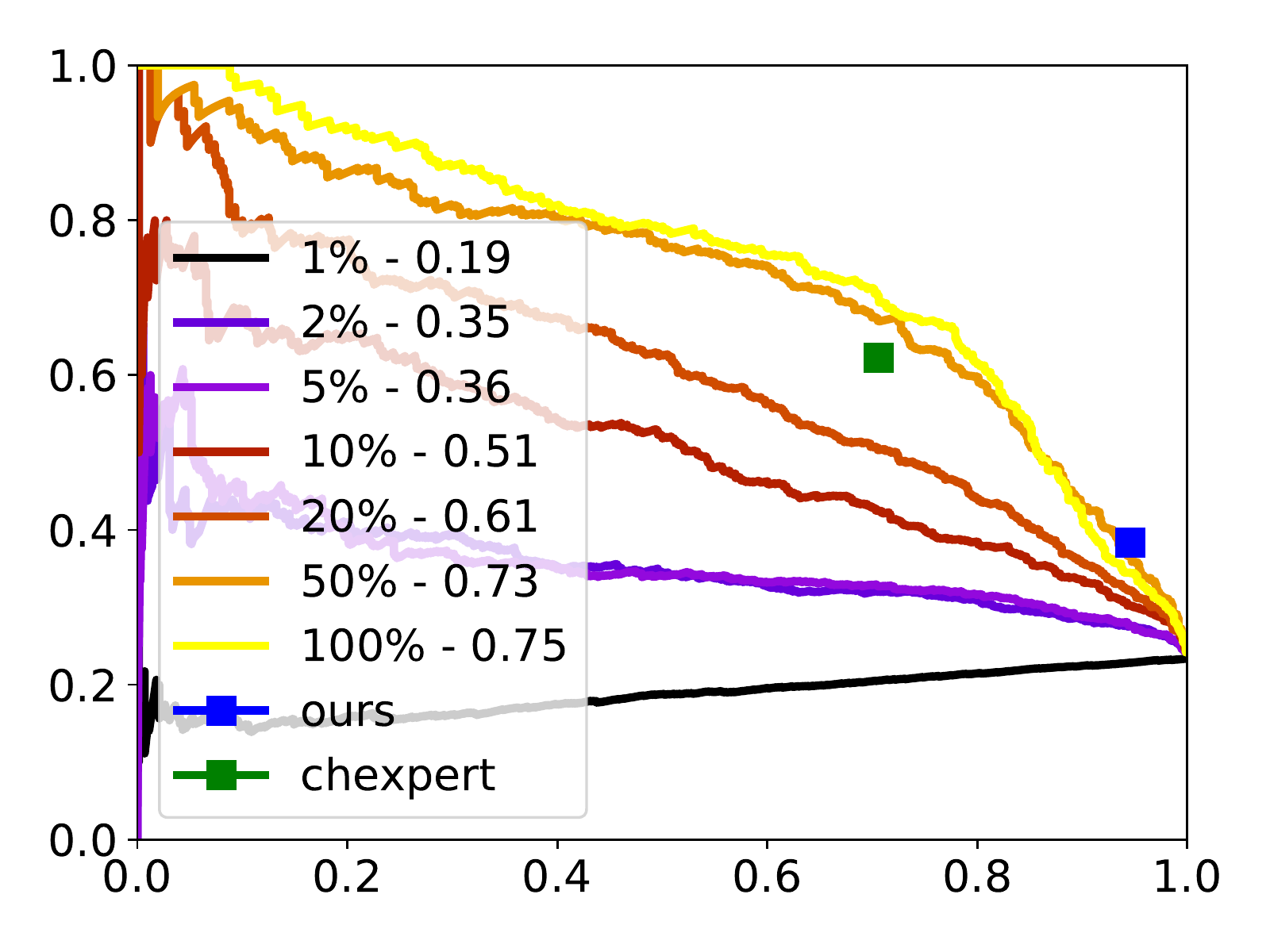}
\caption{Consolidation/Pneumonia}
\label{sfig:rnn_consolidation}
\end{subfigure}%
\begin{subfigure}{.25\linewidth}
\centering
\includegraphics[width=\linewidth]{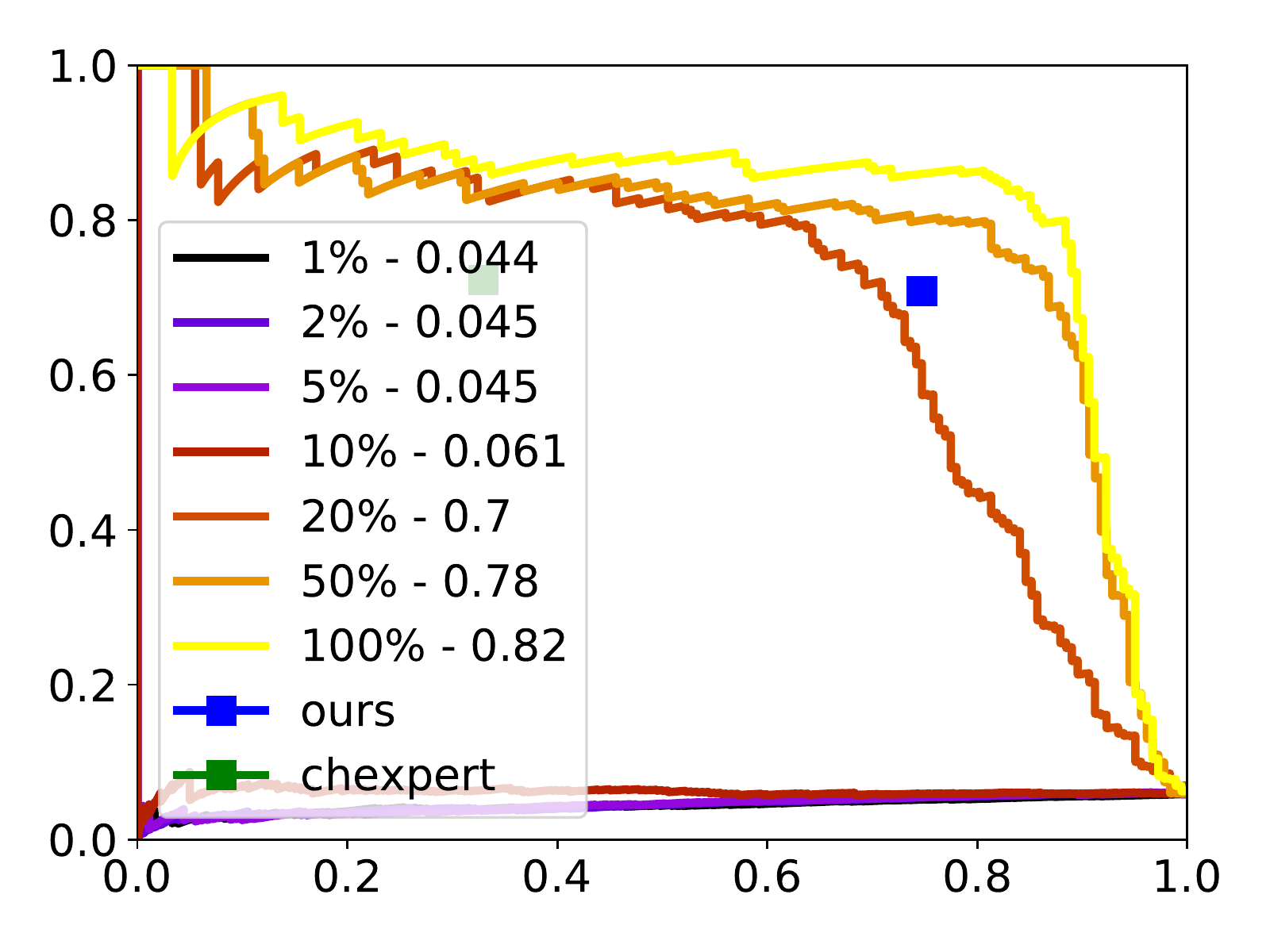}
\caption{Fracture}
\label{sfig:rnn_fracture}
\end{subfigure}%
\begin{subfigure}{.25\textwidth}
\centering
\includegraphics[width=\linewidth]{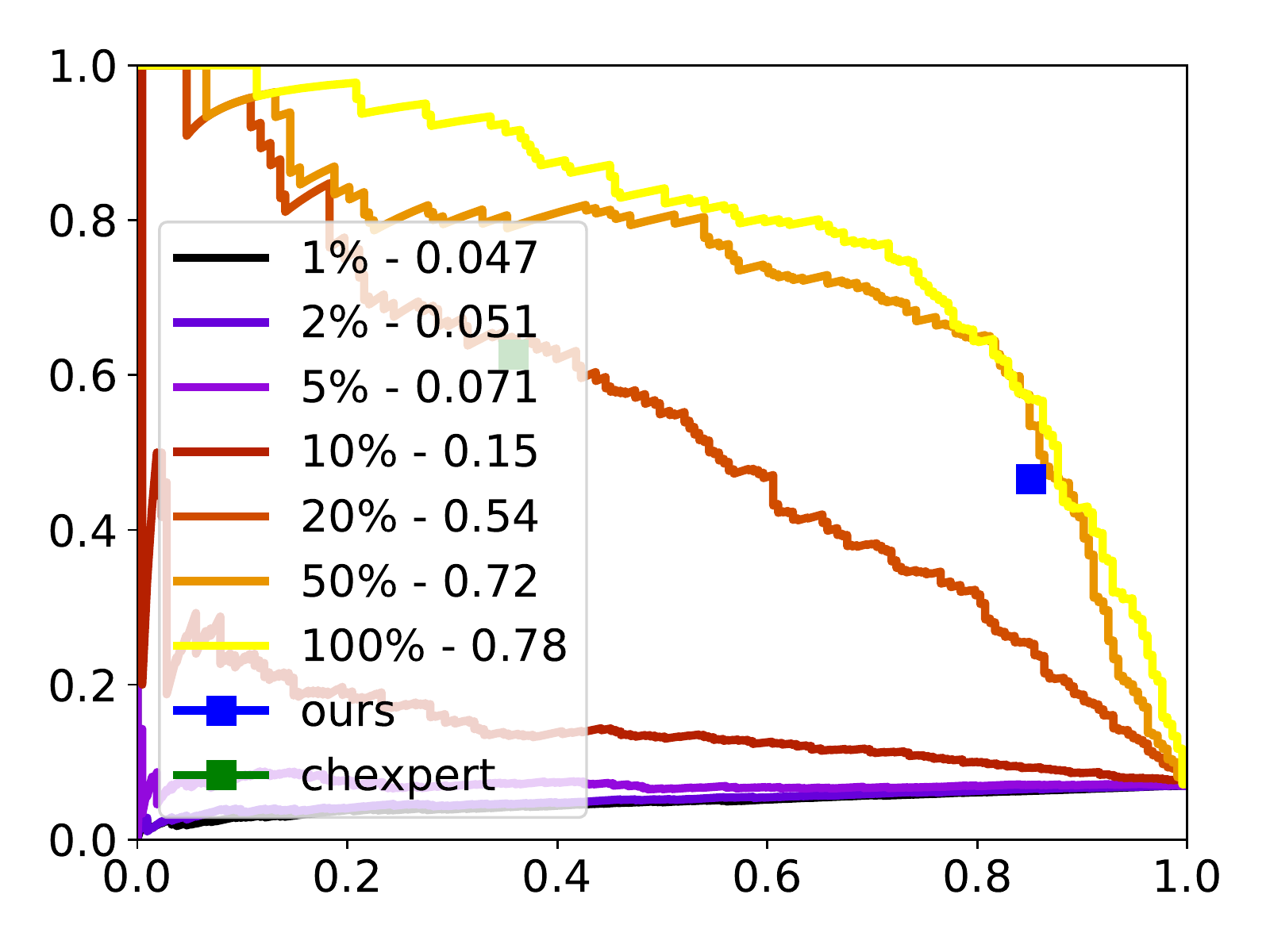}
\caption{Lung Lesion}
\label{sfig:rnn_lesion}
\end{subfigure}\par\medskip
\begin{subfigure}{.25\textwidth}
\centering
\includegraphics[width=\linewidth]{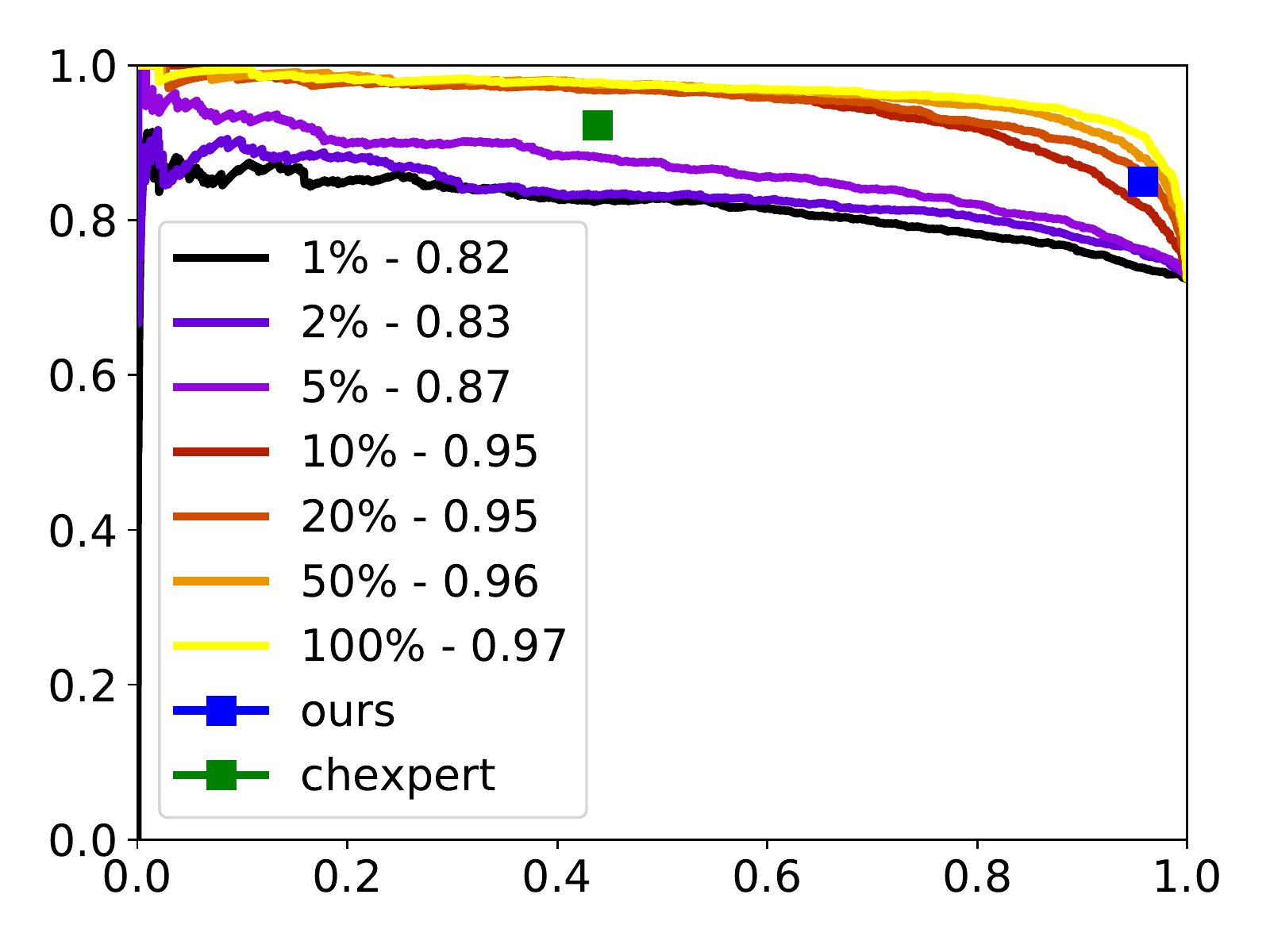}
\caption{Lung Opacity}
\label{sfig:rnn_opacity}
\end{subfigure}%
\begin{subfigure}{.25\linewidth}
\centering
\includegraphics[width=\linewidth]{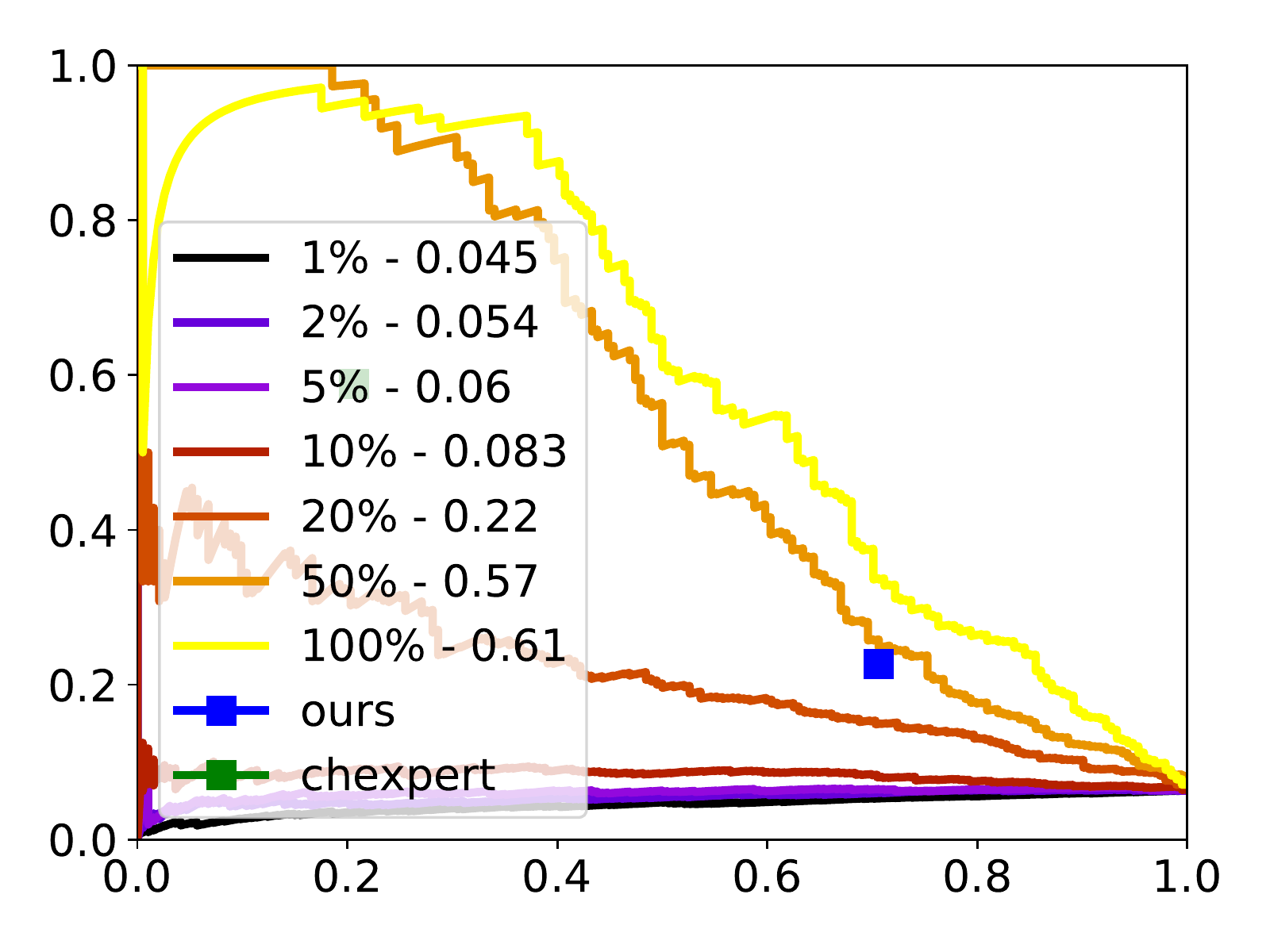}
\caption{Pleural Other}
\label{sfig:rnn_pleural}
\end{subfigure}%
\begin{subfigure}{.25\linewidth}
\centering
\includegraphics[width=\linewidth]{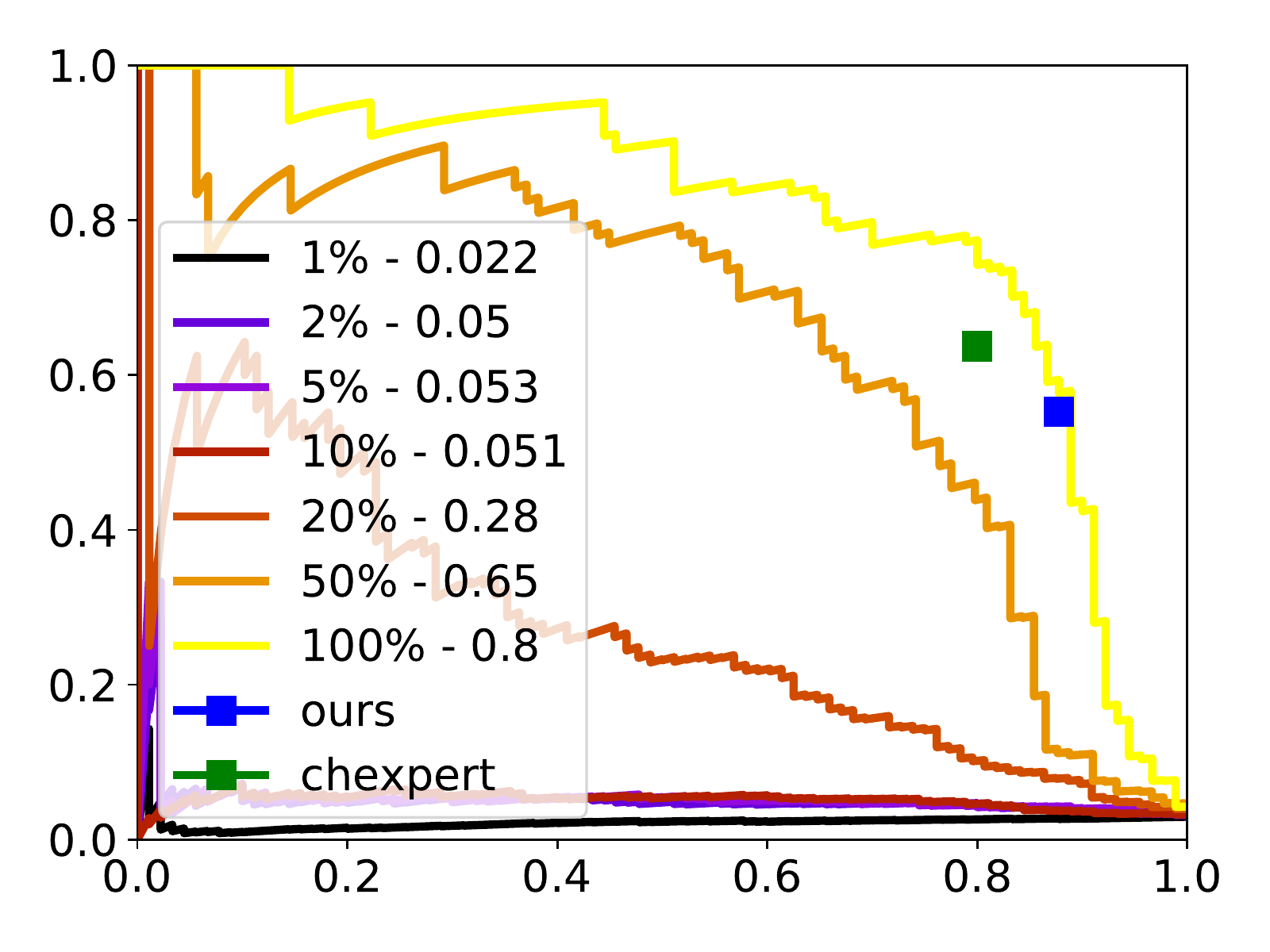}
\caption{Pneumothorax}
\label{sfig:rnn_pneumothorax}
\end{subfigure}%
\begin{subfigure}{.25\textwidth}
\centering
\includegraphics[width=\linewidth]{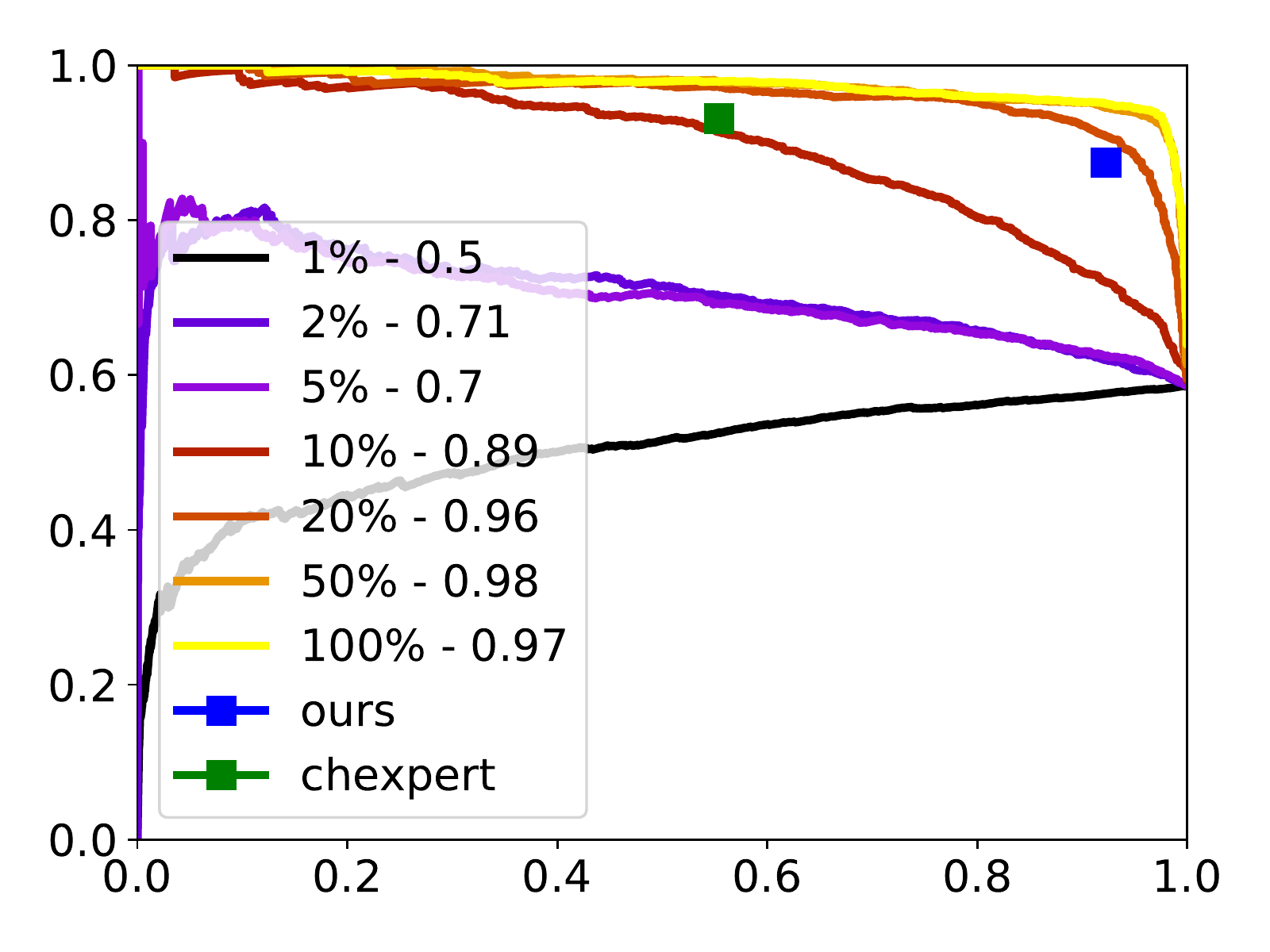}
\caption{Support Devices}
\label{sfig:rnn_support}
\end{subfigure}\par\medskip

\caption{LSTM per-label PR curves on MIMIC test set with different data sizes and comparison with rule-based performance. As commonly shown, the y axis is Precision while the x axis is Recall. We can clearly see the distinction between easy labels and more difficult ones. However, the easiness depends not only on the label itself, but also on how many positive examples exist in the dataset. Finally, we can also notice that with 20\% of the data, the LSTM model outperforms the rule-based baseline for most labels.}
\label{fig:rnn_per_label}
\end{figure*}

\begin{figure*}
\centering
\begin{subfigure}{.25\textwidth}
\centering
\includegraphics[width=\linewidth]{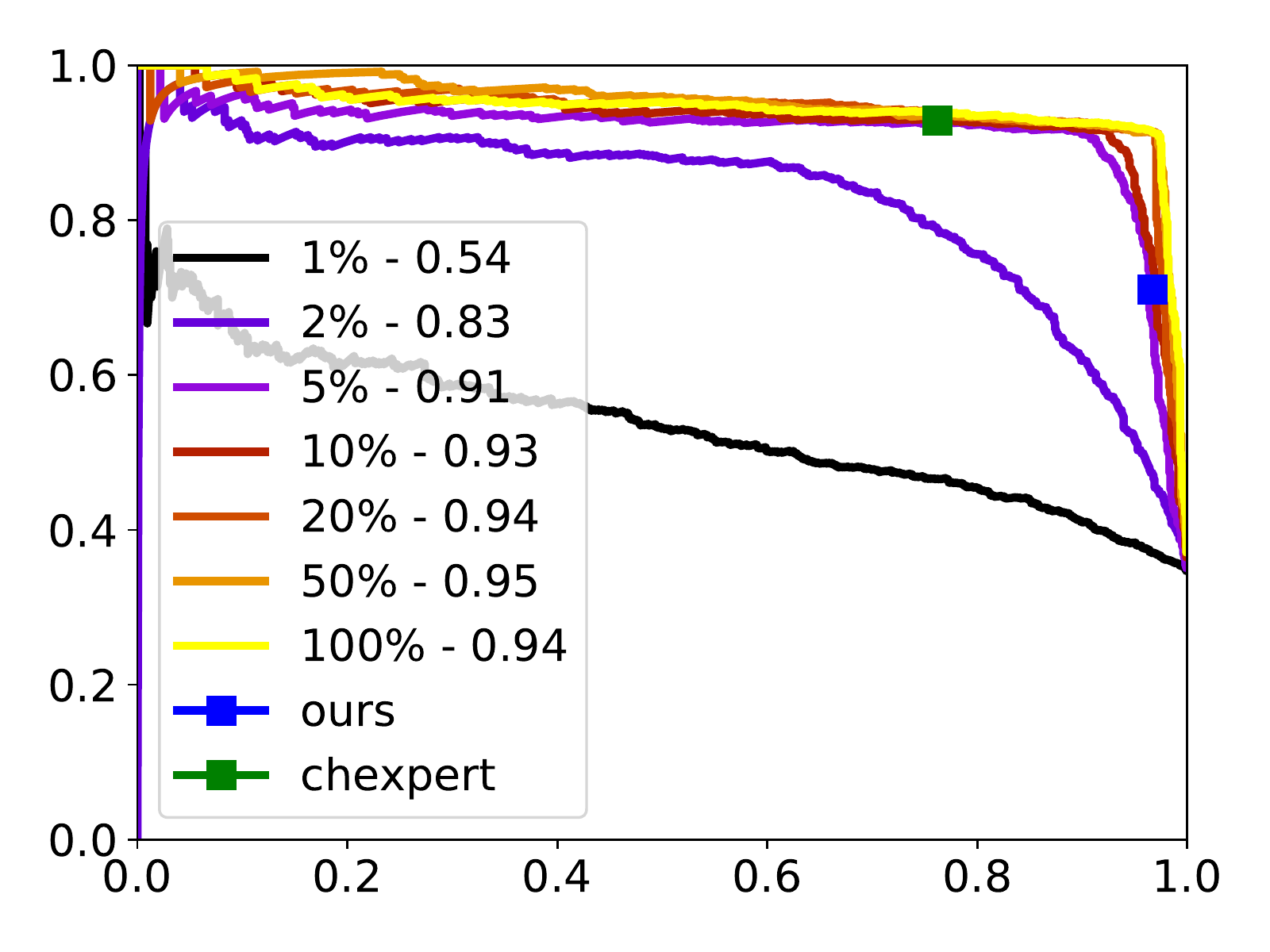}
\caption{Atelectasis}
\label{sfig:cnn_atelectasis}
\end{subfigure}%
\begin{subfigure}{.25\textwidth}
\centering
\includegraphics[width=\linewidth]{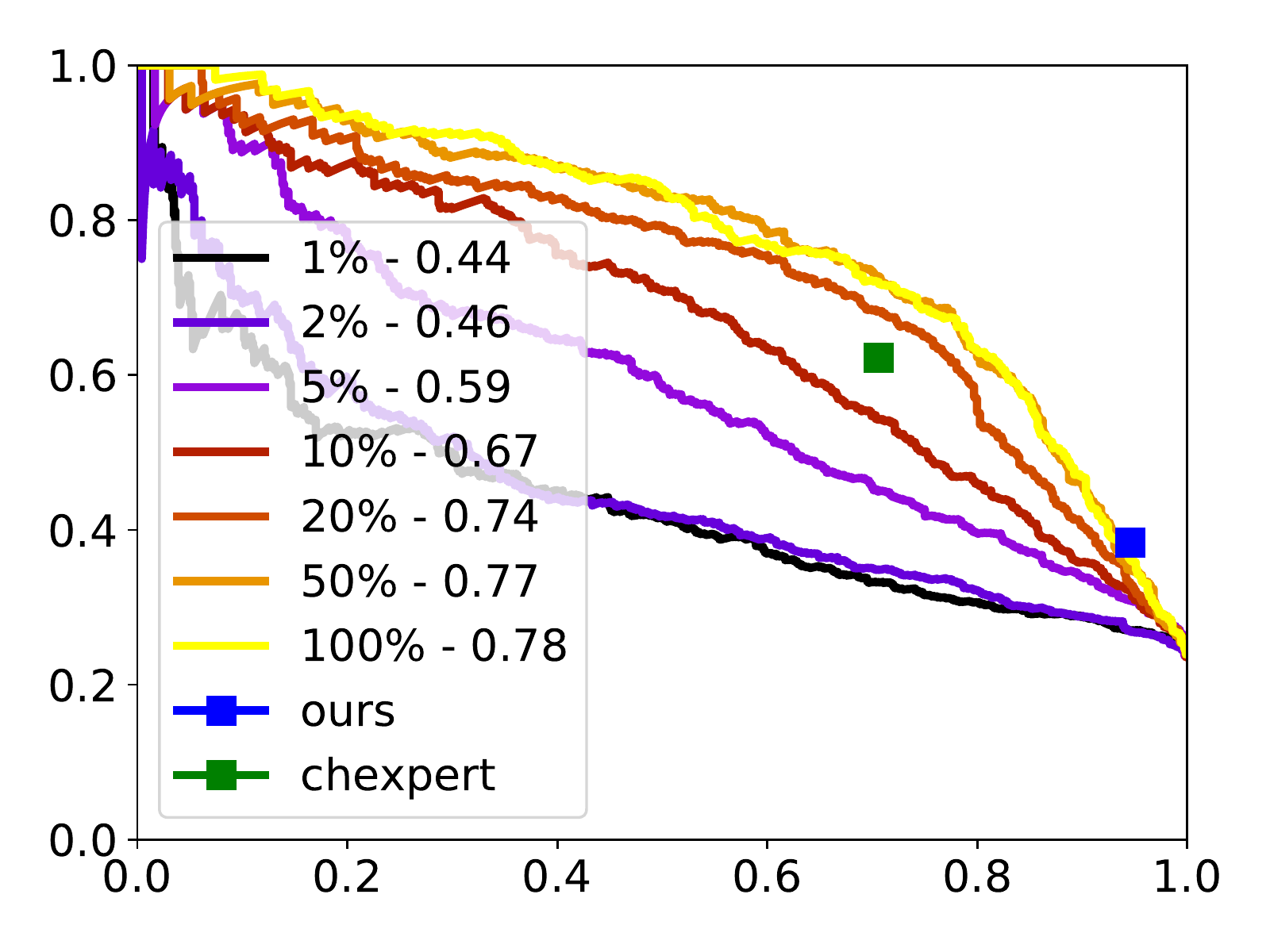}
\caption{Consolidation/Pneumonia}
\label{sfig:cnn_consolidation}
\end{subfigure}%
\begin{subfigure}{.25\linewidth}
\centering
\includegraphics[width=\linewidth]{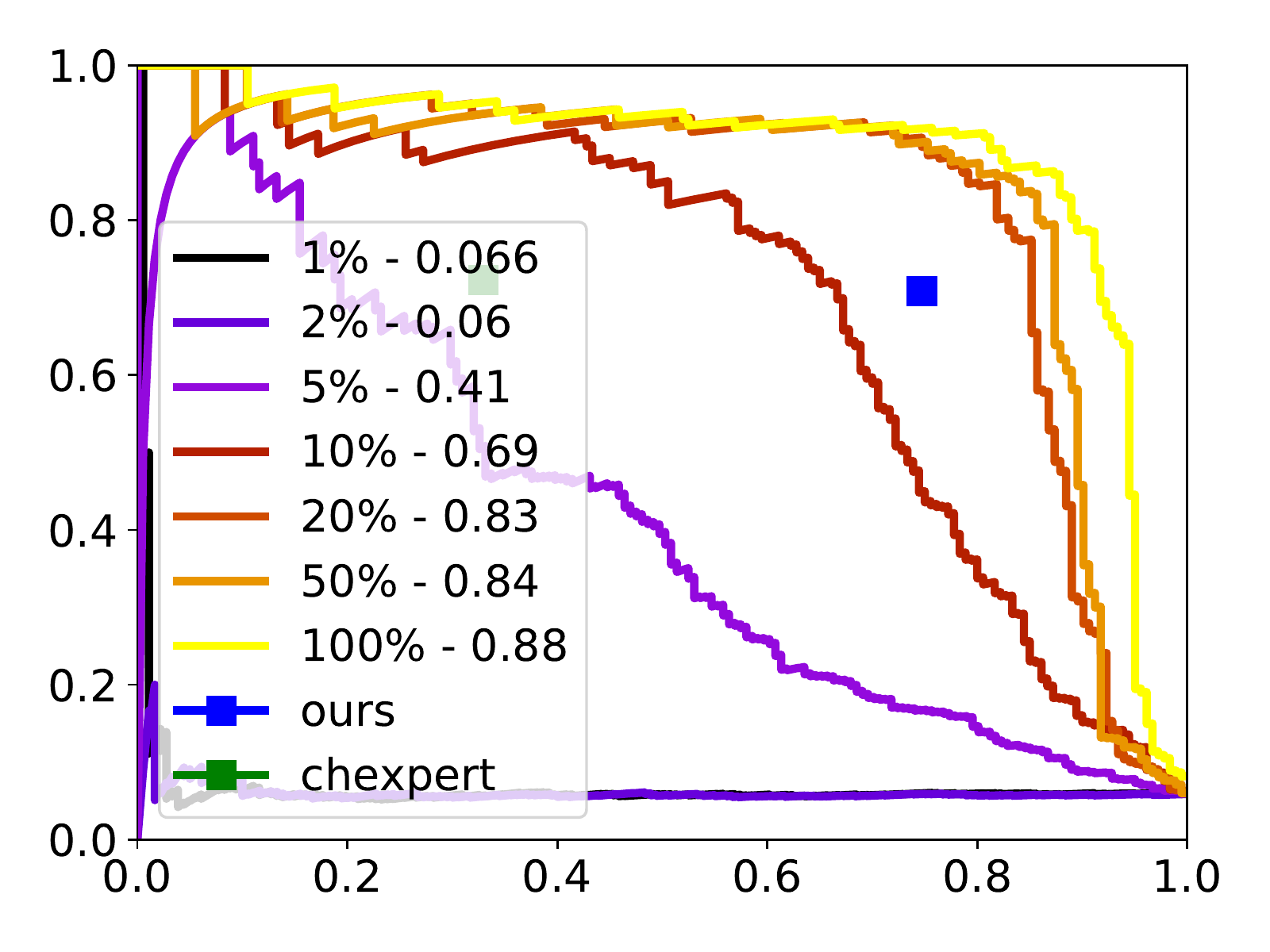}
\caption{Fracture}
\label{sfig:cnn_fracture}
\end{subfigure}%
\begin{subfigure}{.25\textwidth}
\centering
\includegraphics[width=\linewidth]{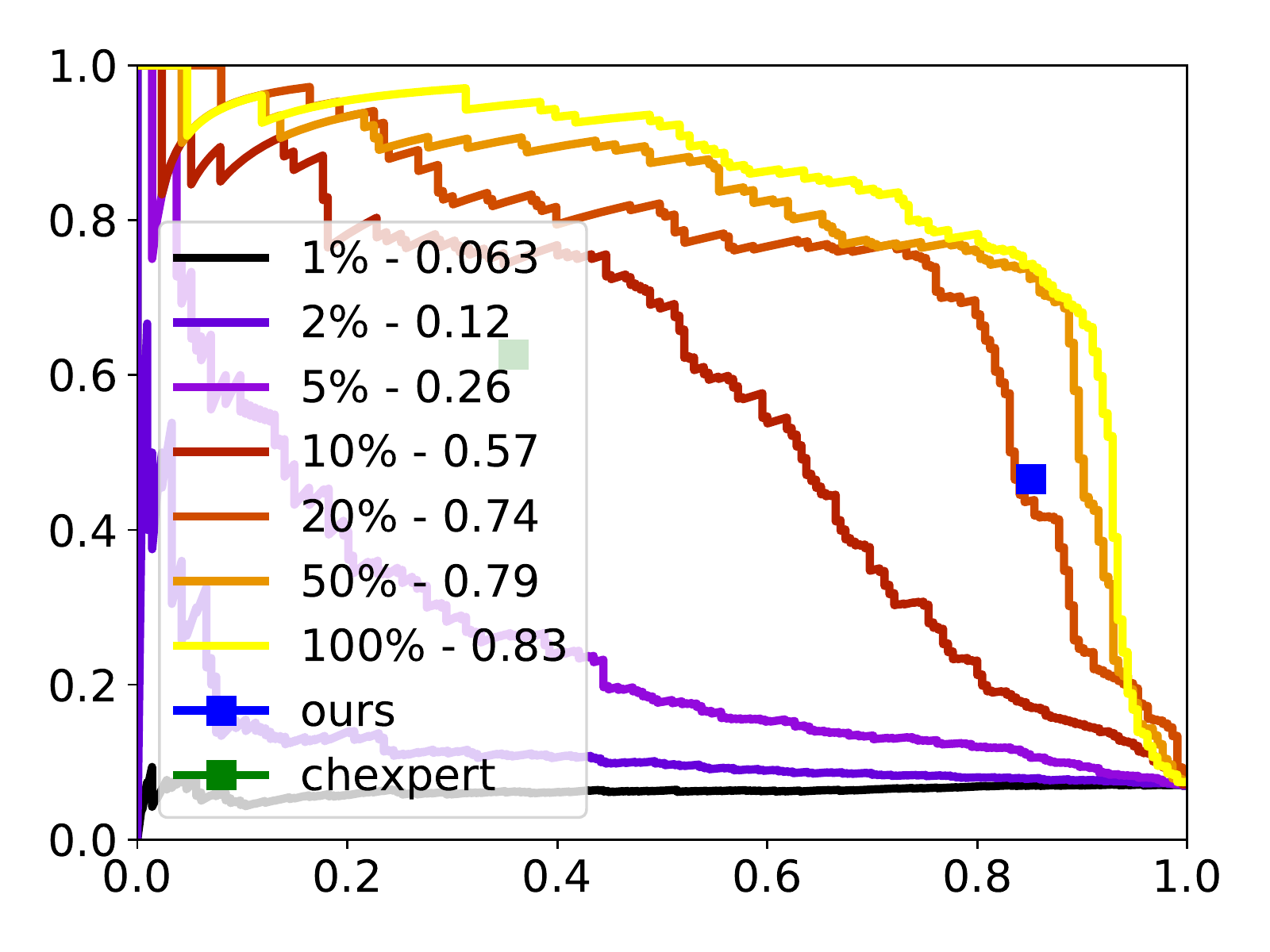}
\caption{Lung Lesion}
\label{sfig:cnn_lesion}
\end{subfigure}\par\medskip
\begin{subfigure}{.25\textwidth}
\centering
\includegraphics[width=\linewidth]{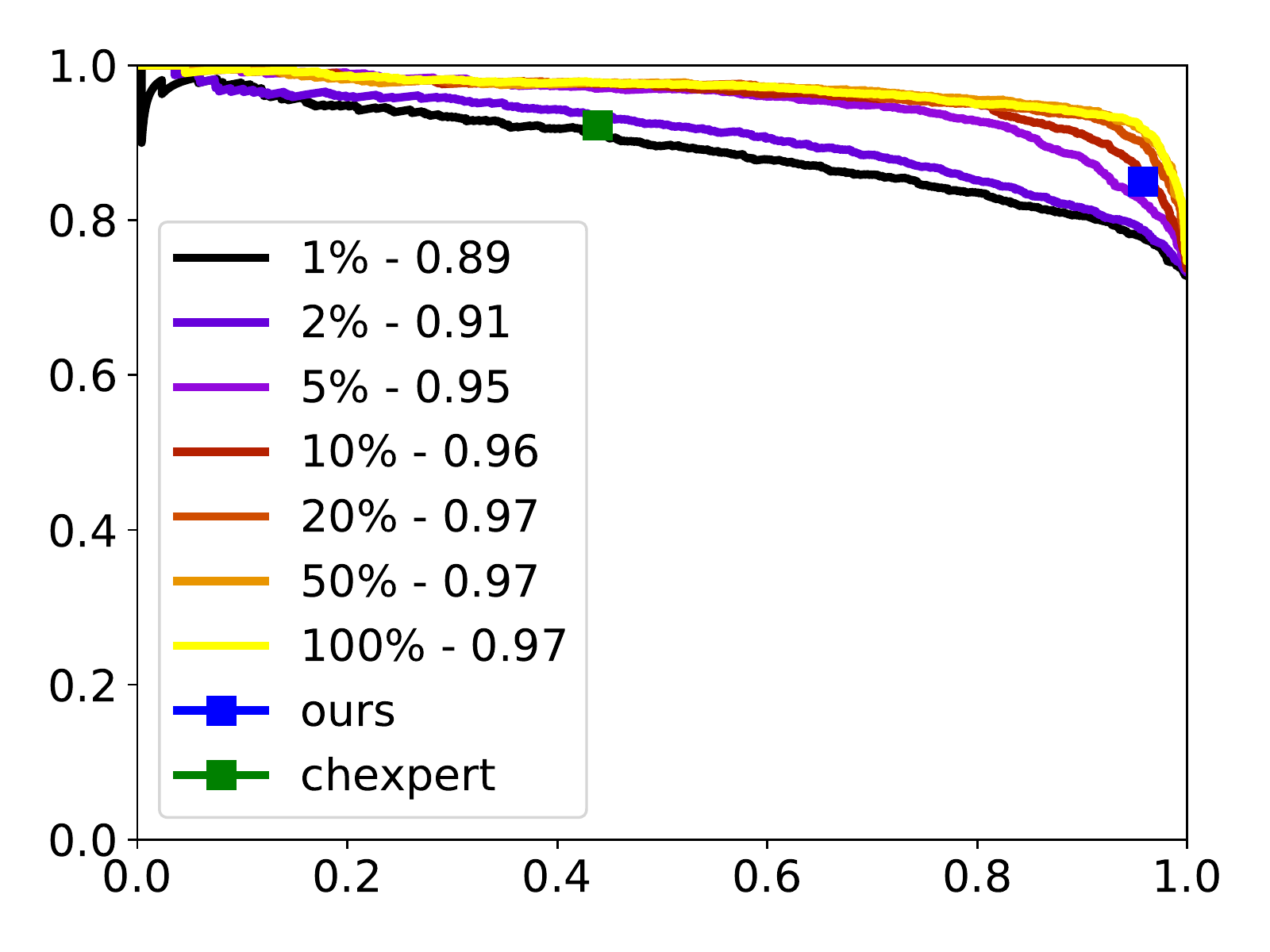}
\caption{Lung Opacity}
\label{sfig:cnn_opacity}
\end{subfigure}%
\begin{subfigure}{.25\linewidth}
\centering
\includegraphics[width=\linewidth]{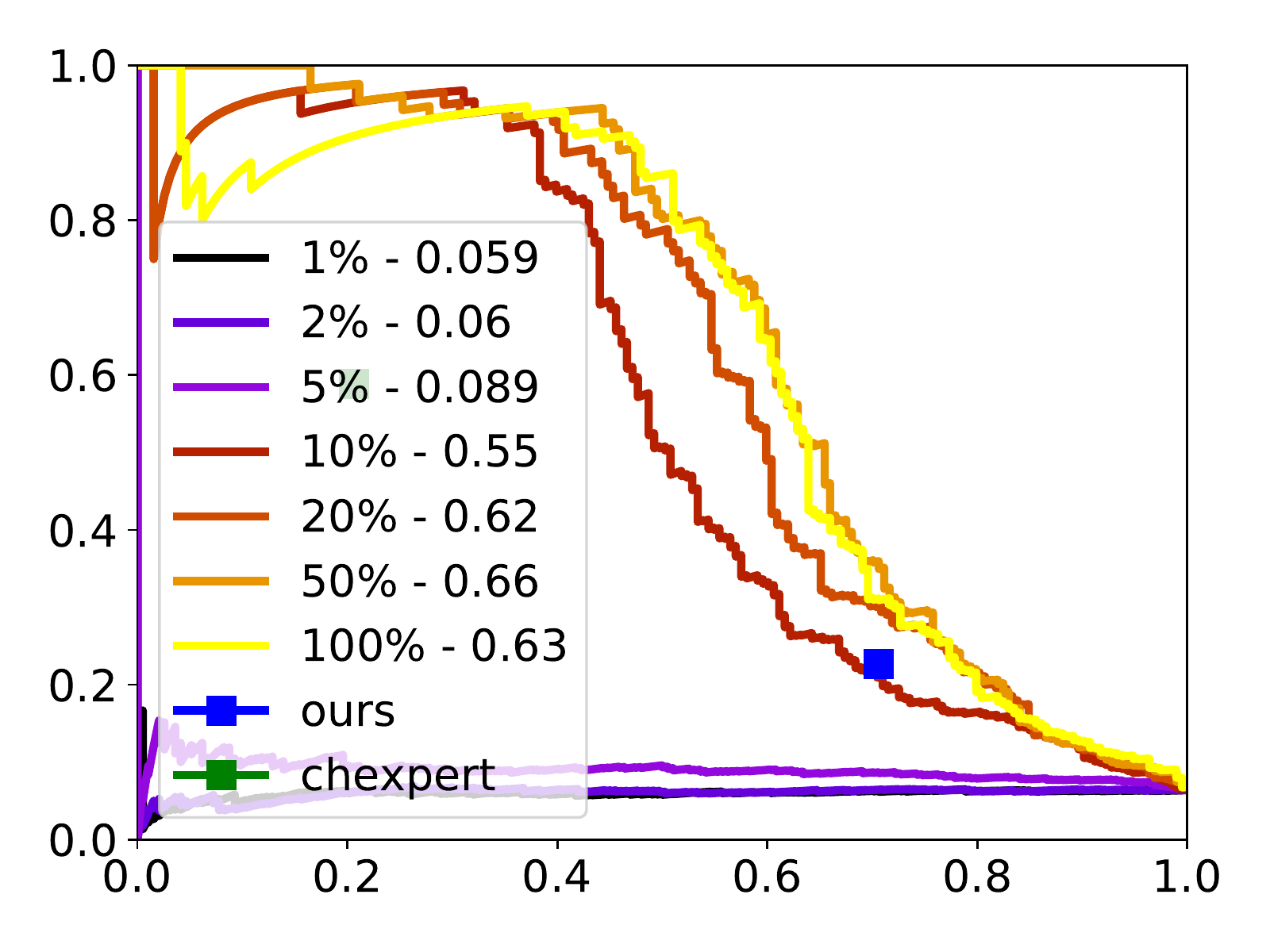}
\caption{Pleural Other}
\label{sfig:cnn_pleural}
\end{subfigure}%
\begin{subfigure}{.25\linewidth}
\centering
\includegraphics[width=\linewidth]{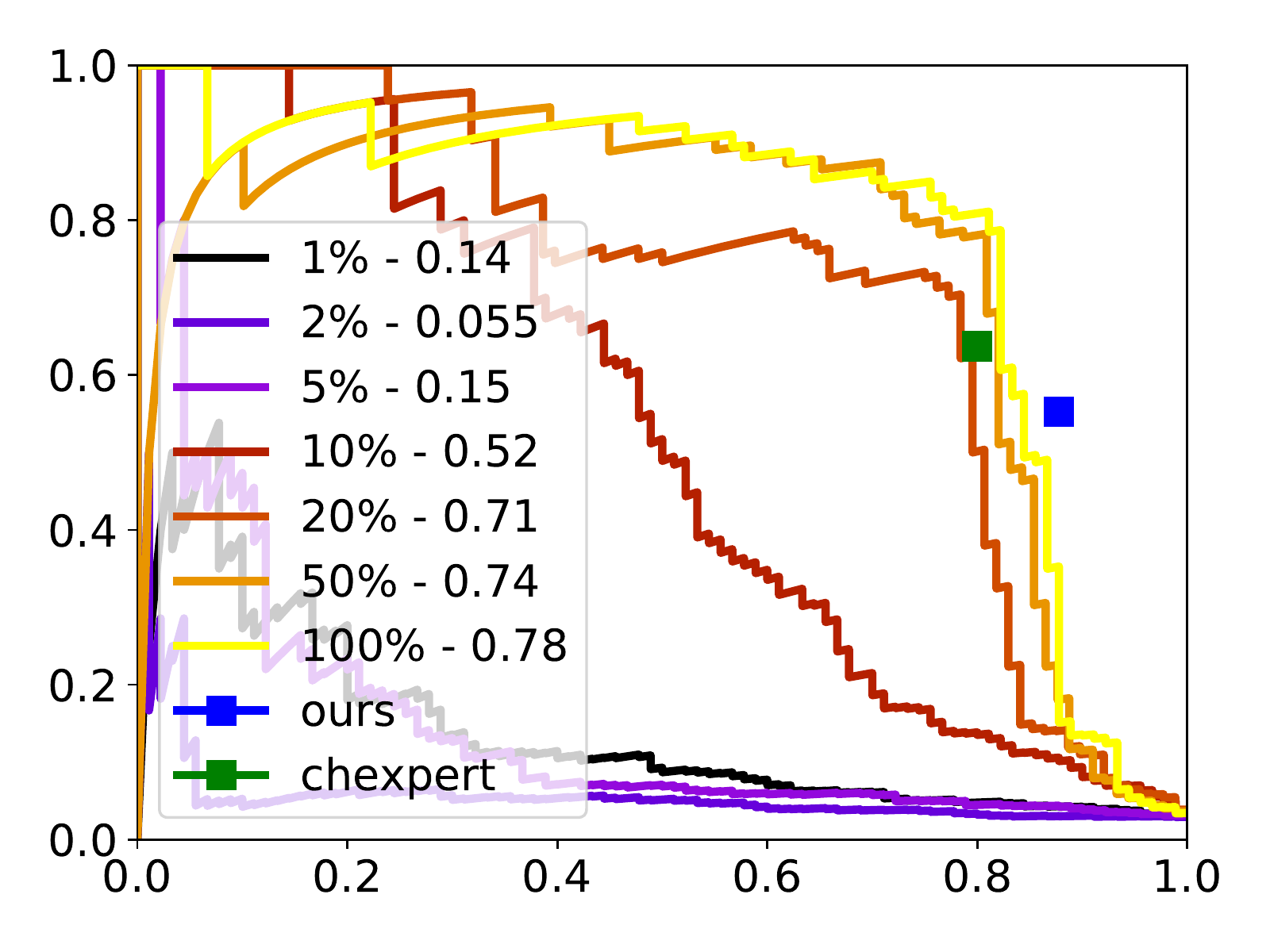}
\caption{Pneumothorax}
\label{sfig:cnn_pneumothorax}
\end{subfigure}%
\begin{subfigure}{.25\textwidth}
\centering
\includegraphics[width=\linewidth]{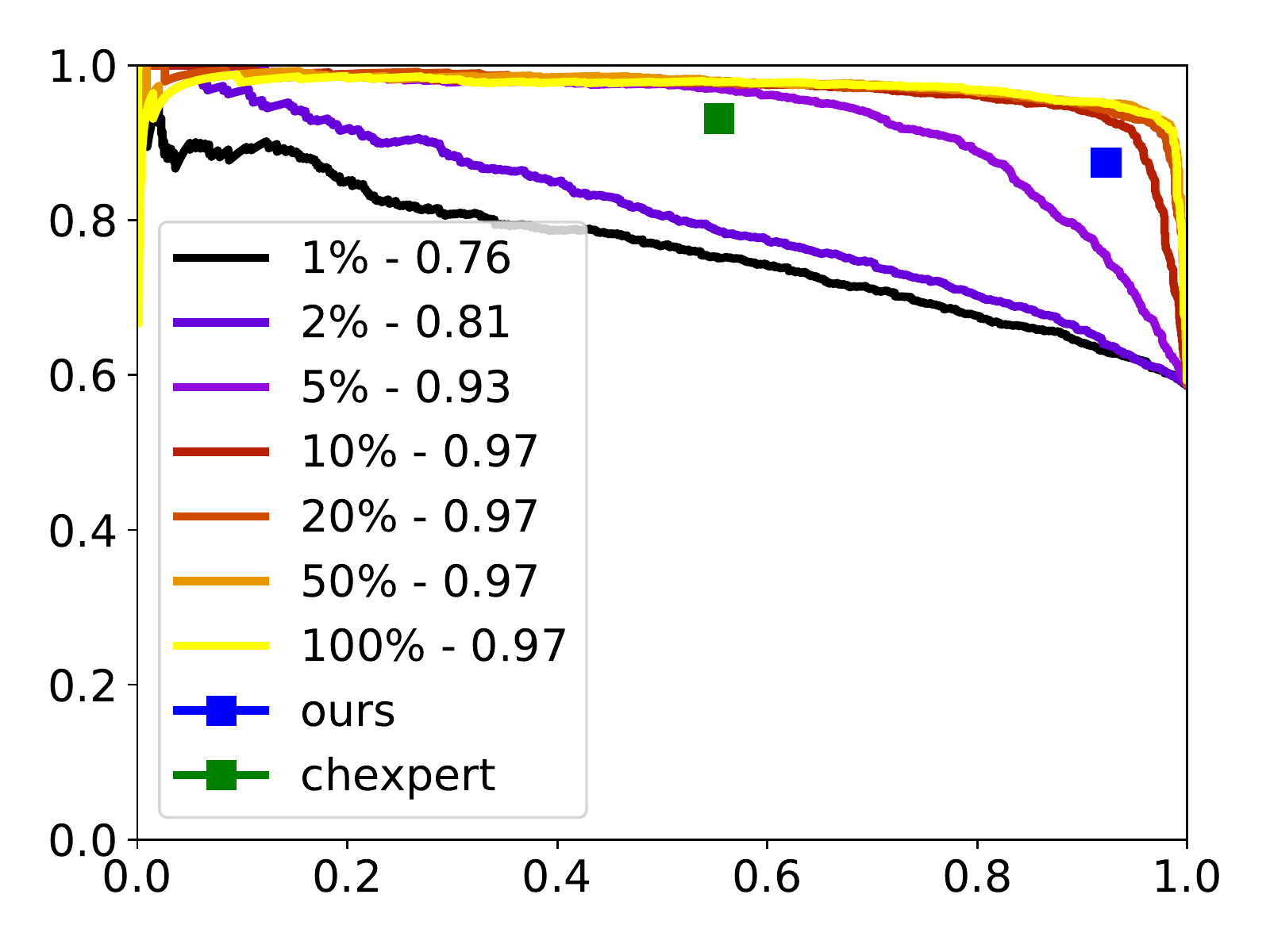}
\caption{Support Devices}
\label{sfig:cnn_support}
\end{subfigure}\par\medskip

\caption{CNN per-label PR curves on MIMIC test set with different data sizes and comparison with rule-based performance. As commonly shown, the y axis is Precision while the x axis is Recall. The results are similar to the LSTM, except that the CNN model seems to be able to learn from a small amount of data with more confidence. This is probably due to the ability of the model to focus on keywords through the max pooling operation.}
\label{fig:cnn_per_label}
\end{figure*}

\begin{figure*}
\centering
\begin{subfigure}{.25\textwidth}
\centering
\includegraphics[width=\linewidth]{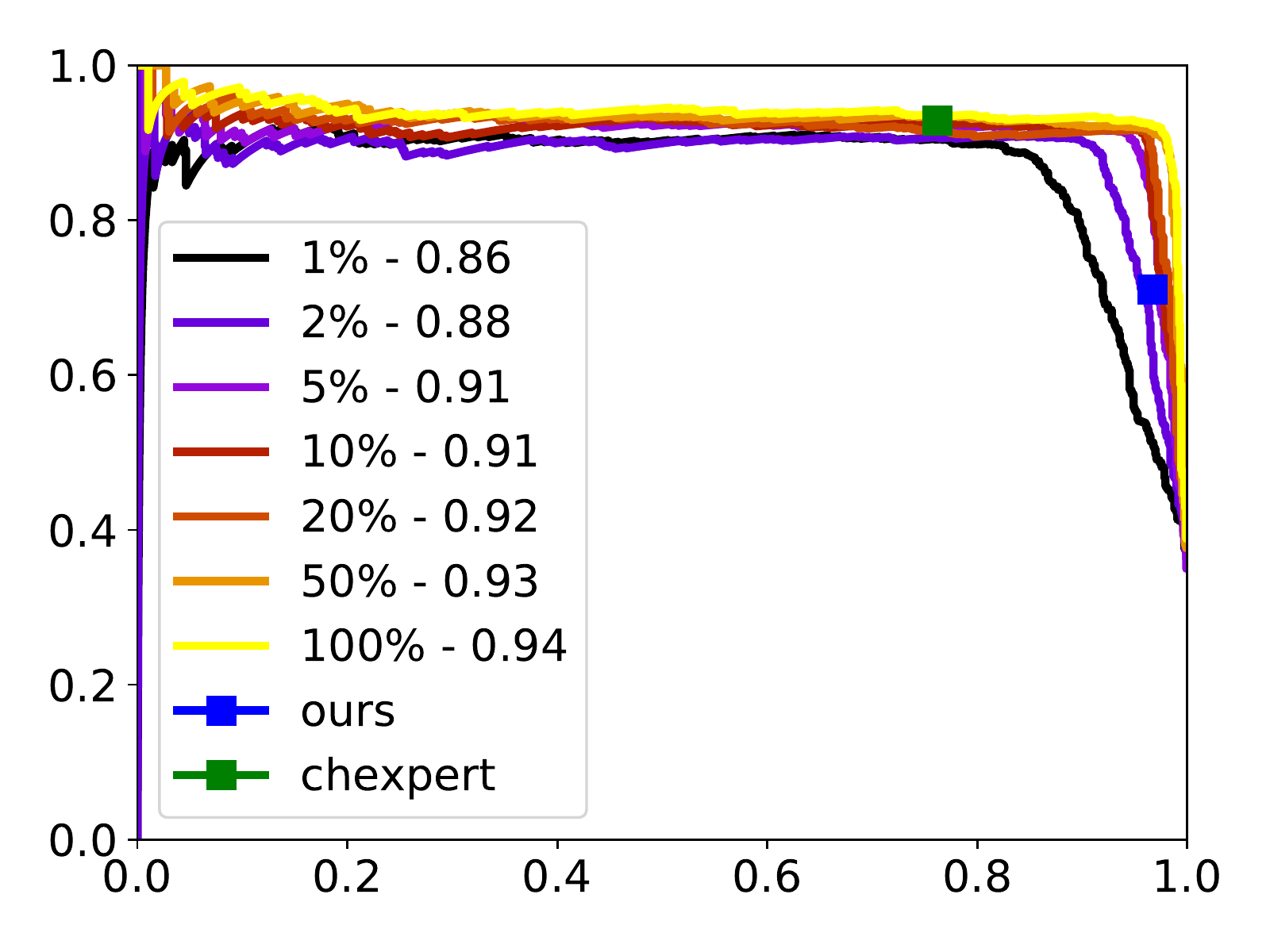}
\caption{Atelectasis}
\label{sfig:rcnn_atelectasis}
\end{subfigure}%
\begin{subfigure}{.25\textwidth}
\centering
\includegraphics[width=\linewidth]{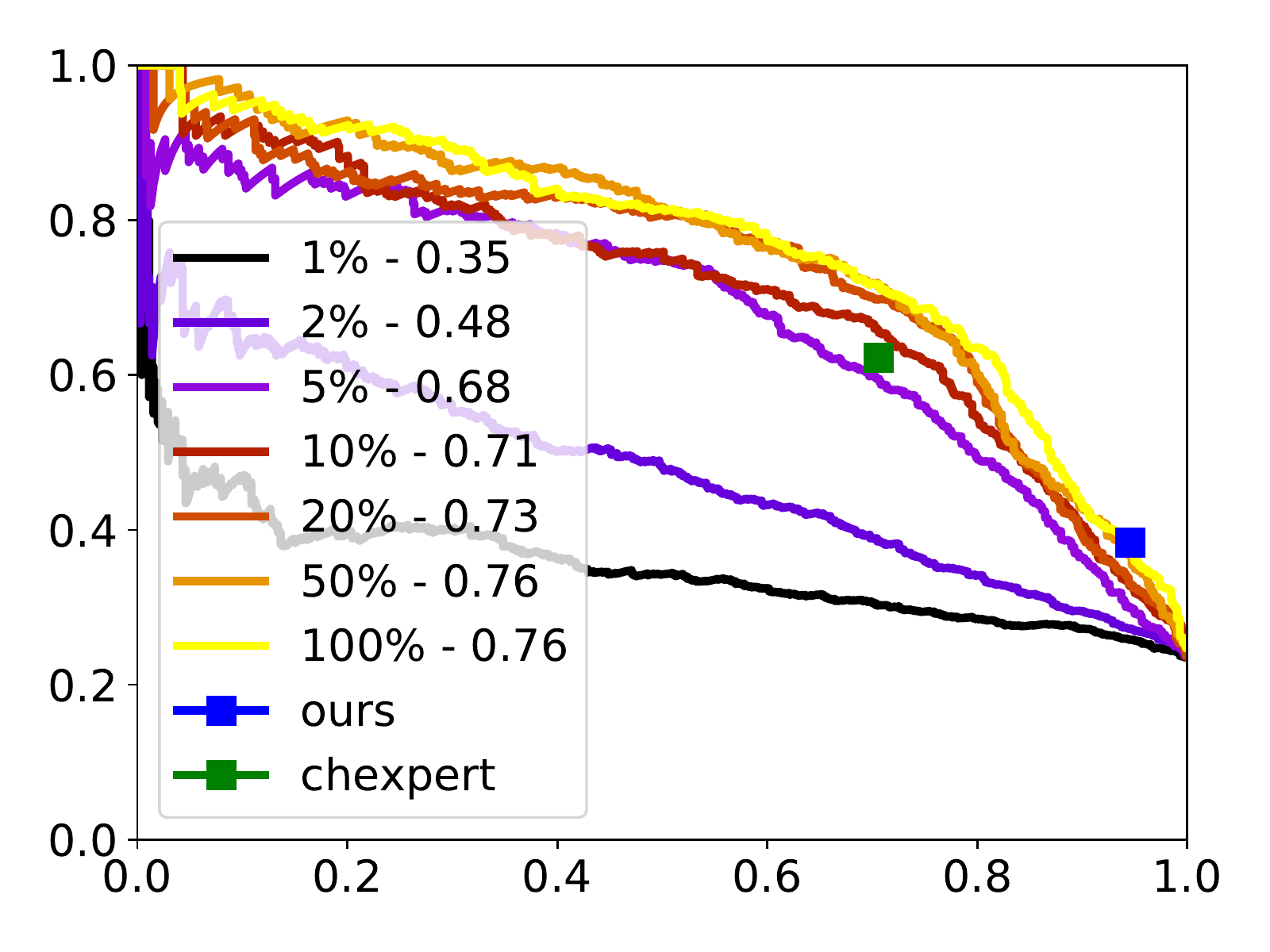}
\caption{Consolidation/Pneumonia}
\label{sfig:rcnn_consolidation}
\end{subfigure}%
\begin{subfigure}{.25\linewidth}
\centering
\includegraphics[width=\linewidth]{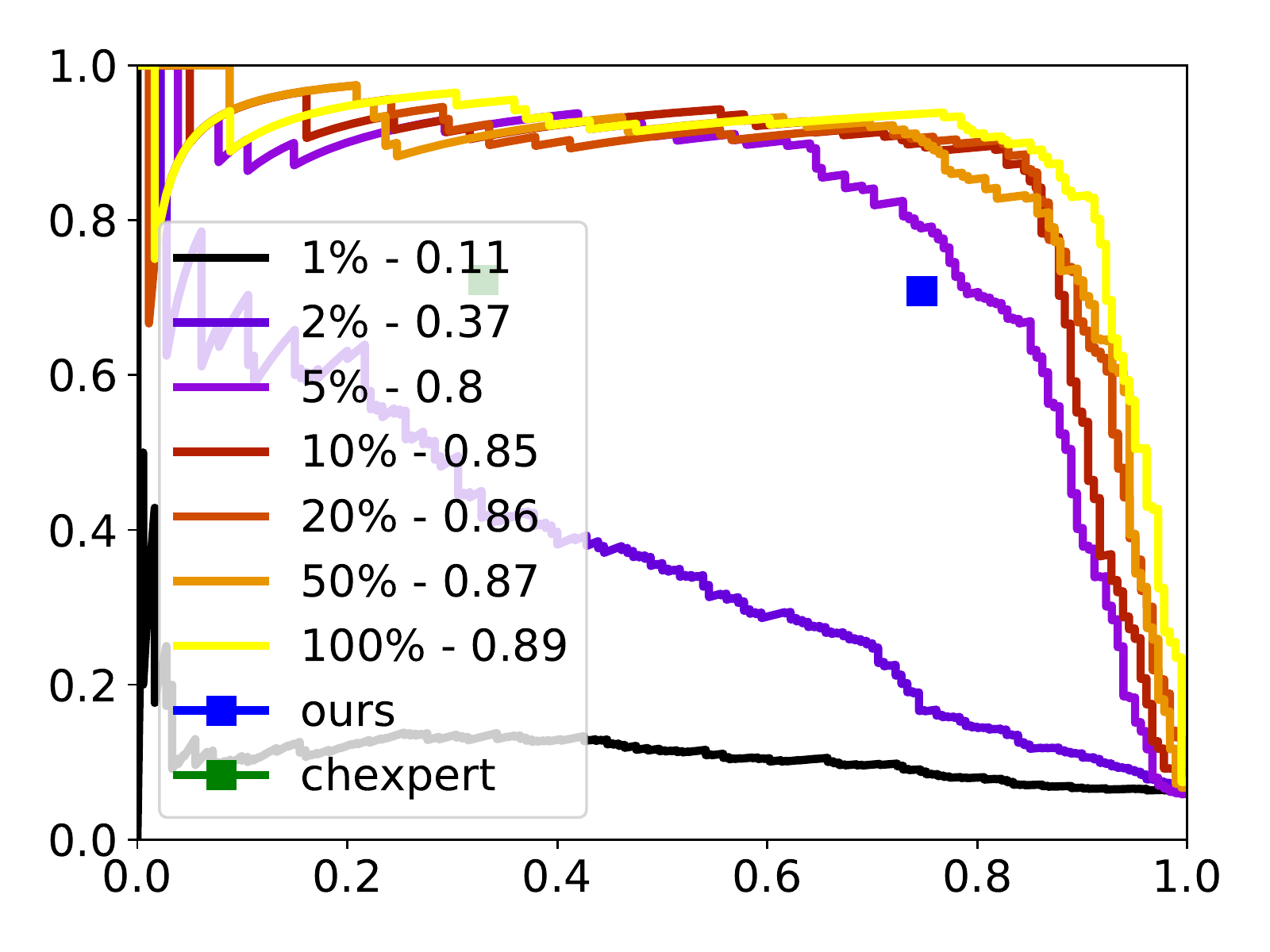}
\caption{Fracture}
\label{sfig:rcnn_fracture}
\end{subfigure}%
\begin{subfigure}{.25\textwidth}
\centering
\includegraphics[width=\linewidth]{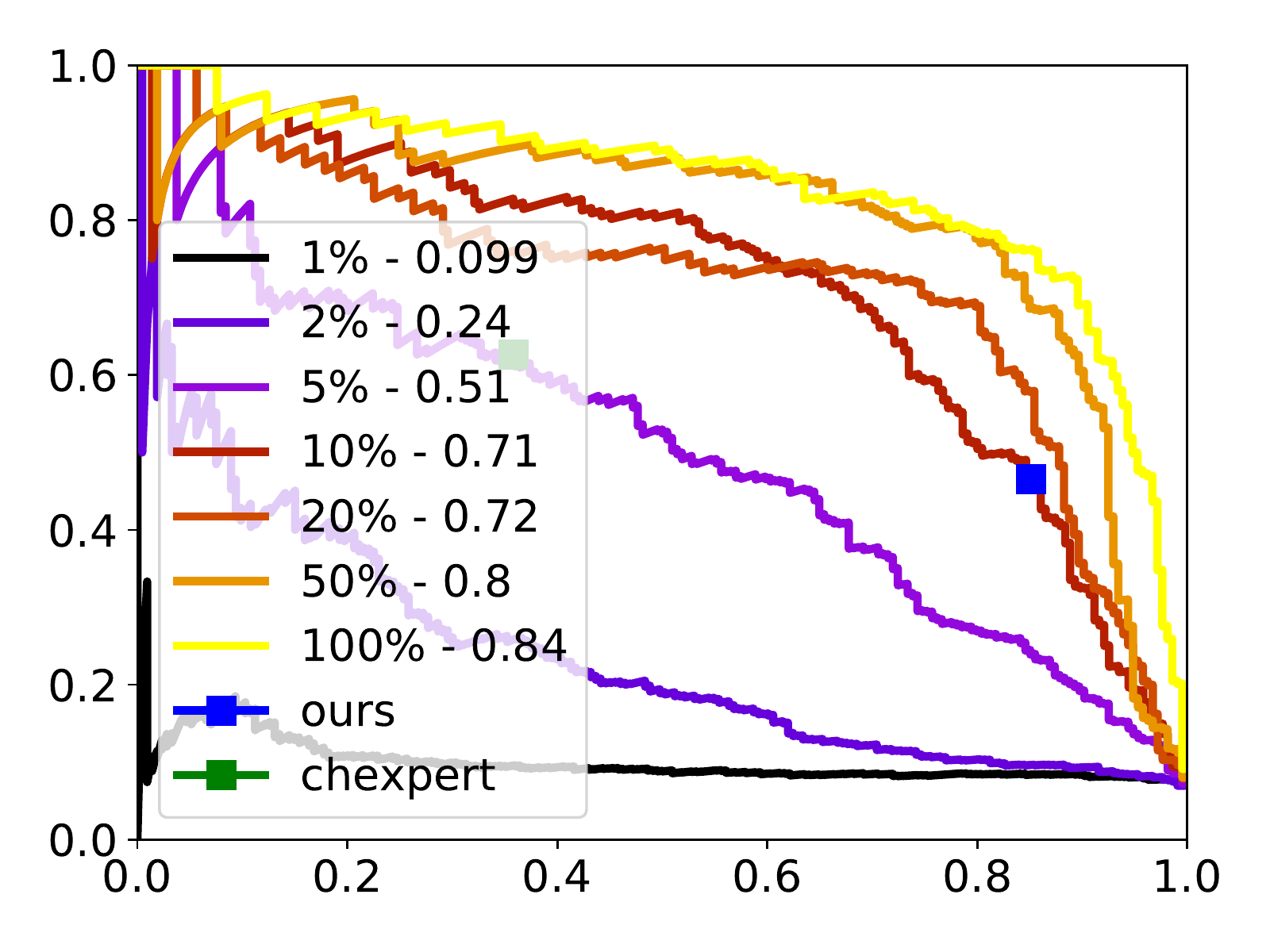}
\caption{Lung Lesion}
\label{sfig:rcnn_lesion}
\end{subfigure}\par\medskip
\begin{subfigure}{.25\textwidth}
\centering
\includegraphics[width=\linewidth]{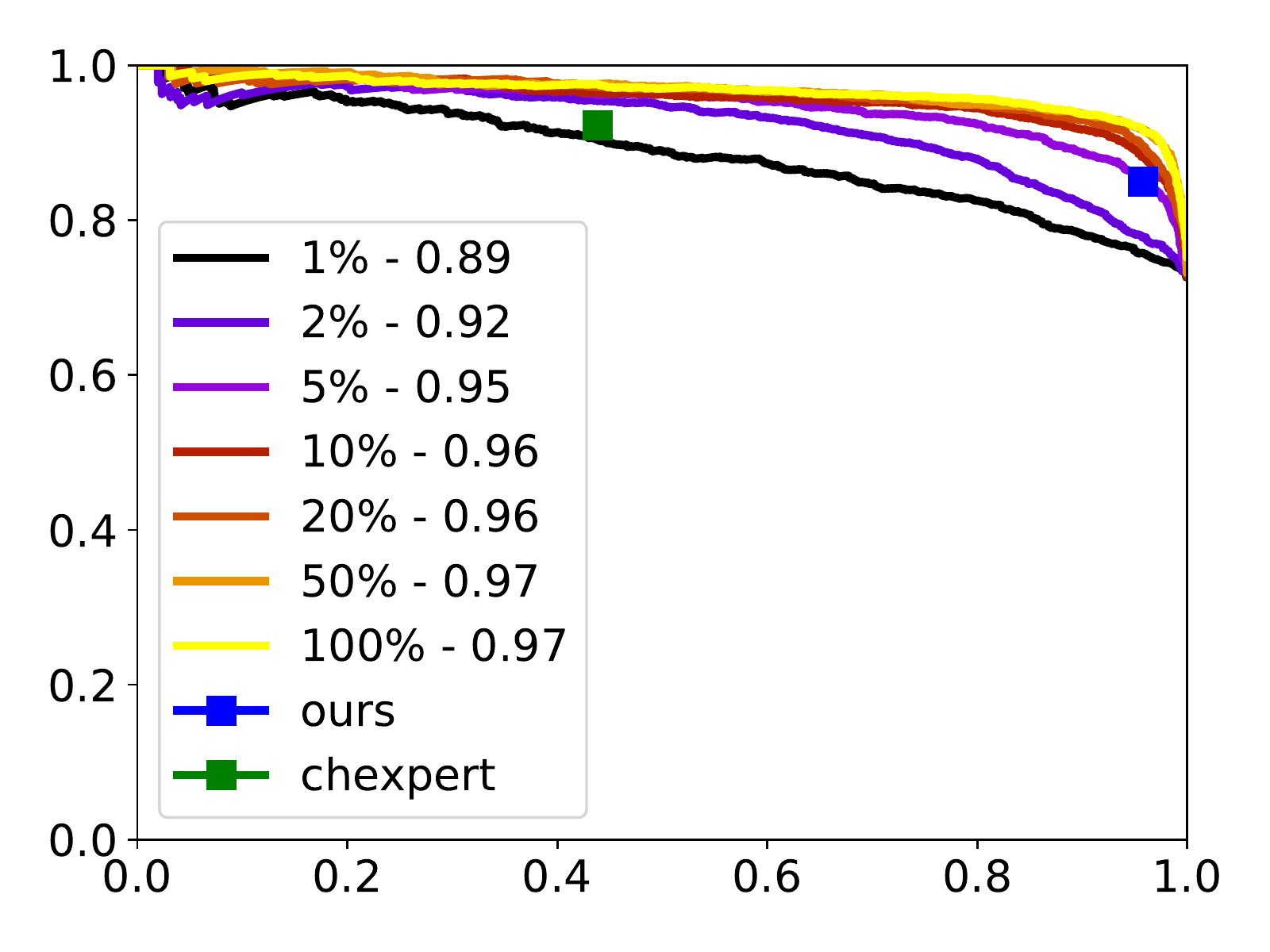}
\caption{Lung Opacity}
\label{sfig:rcnn_opacity}
\end{subfigure}%
\begin{subfigure}{.25\linewidth}
\centering
\includegraphics[width=\linewidth]{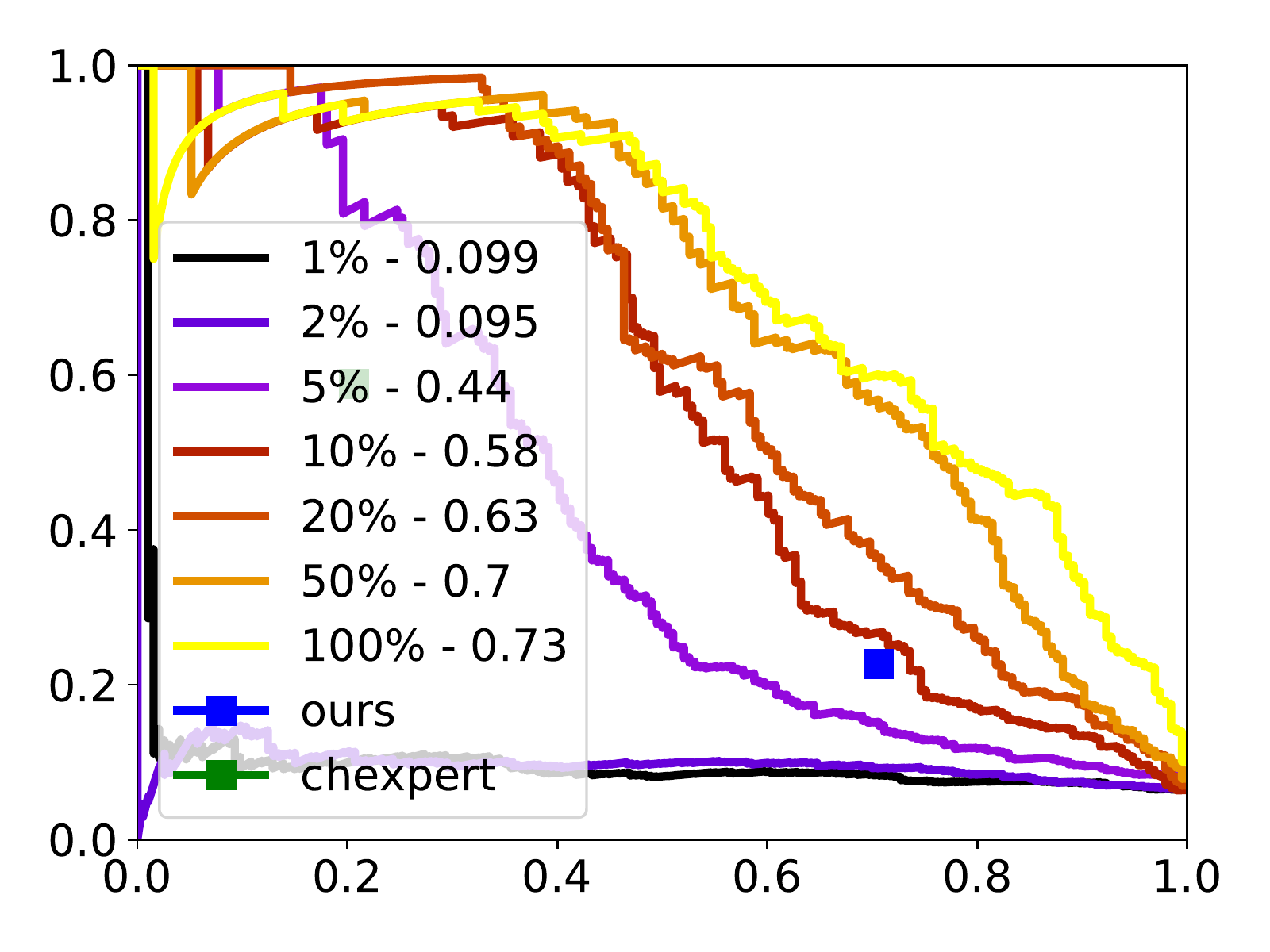}
\caption{Pleural Other}
\label{sfig:rcnn_pleural}
\end{subfigure}%
\begin{subfigure}{.25\linewidth}
\centering
\includegraphics[width=\linewidth]{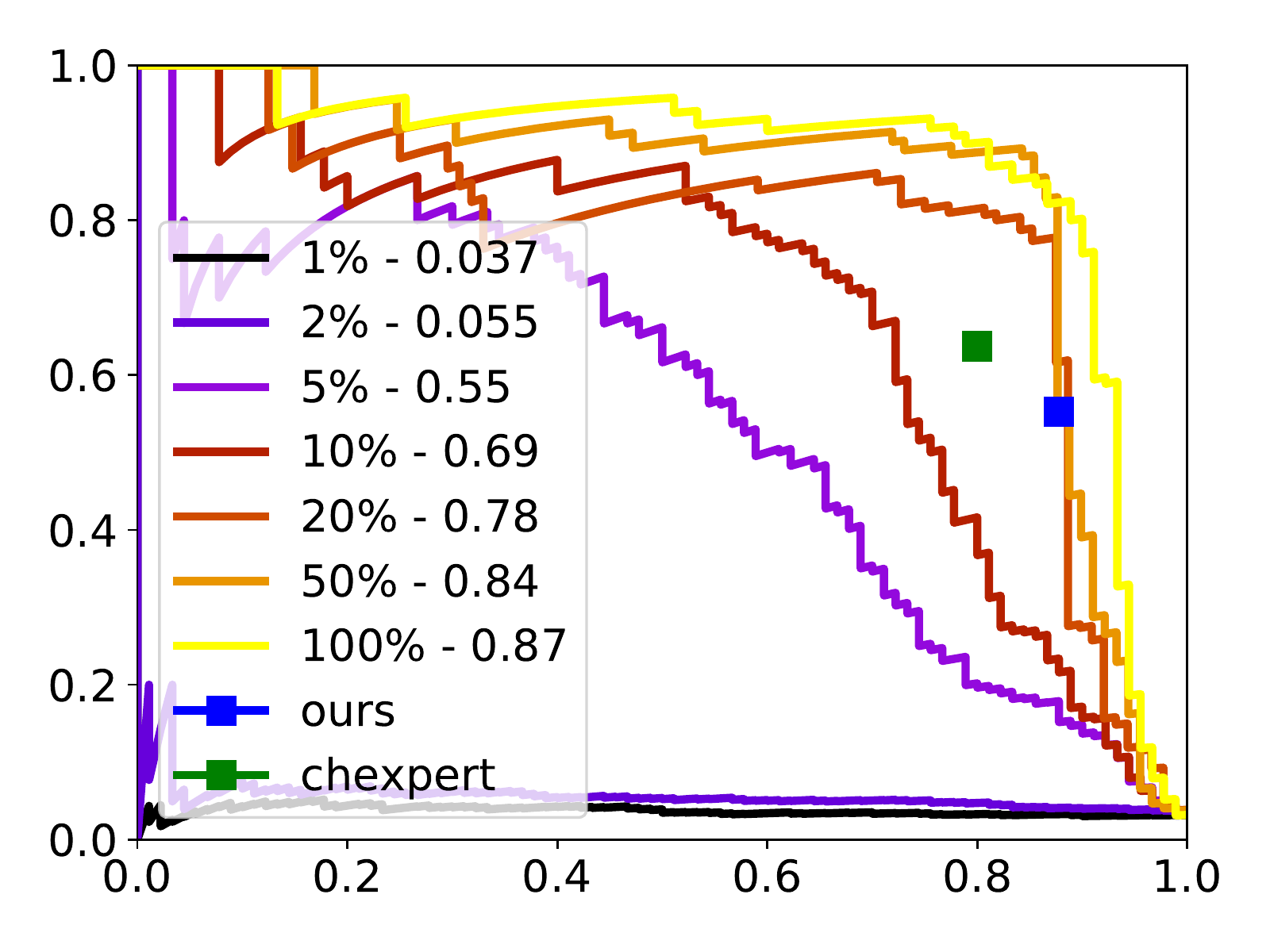}
\caption{Pneumothorax}
\label{sfig:rcnn_pneumothorax}
\end{subfigure}%
\begin{subfigure}{.25\textwidth}
\centering
\includegraphics[width=\linewidth]{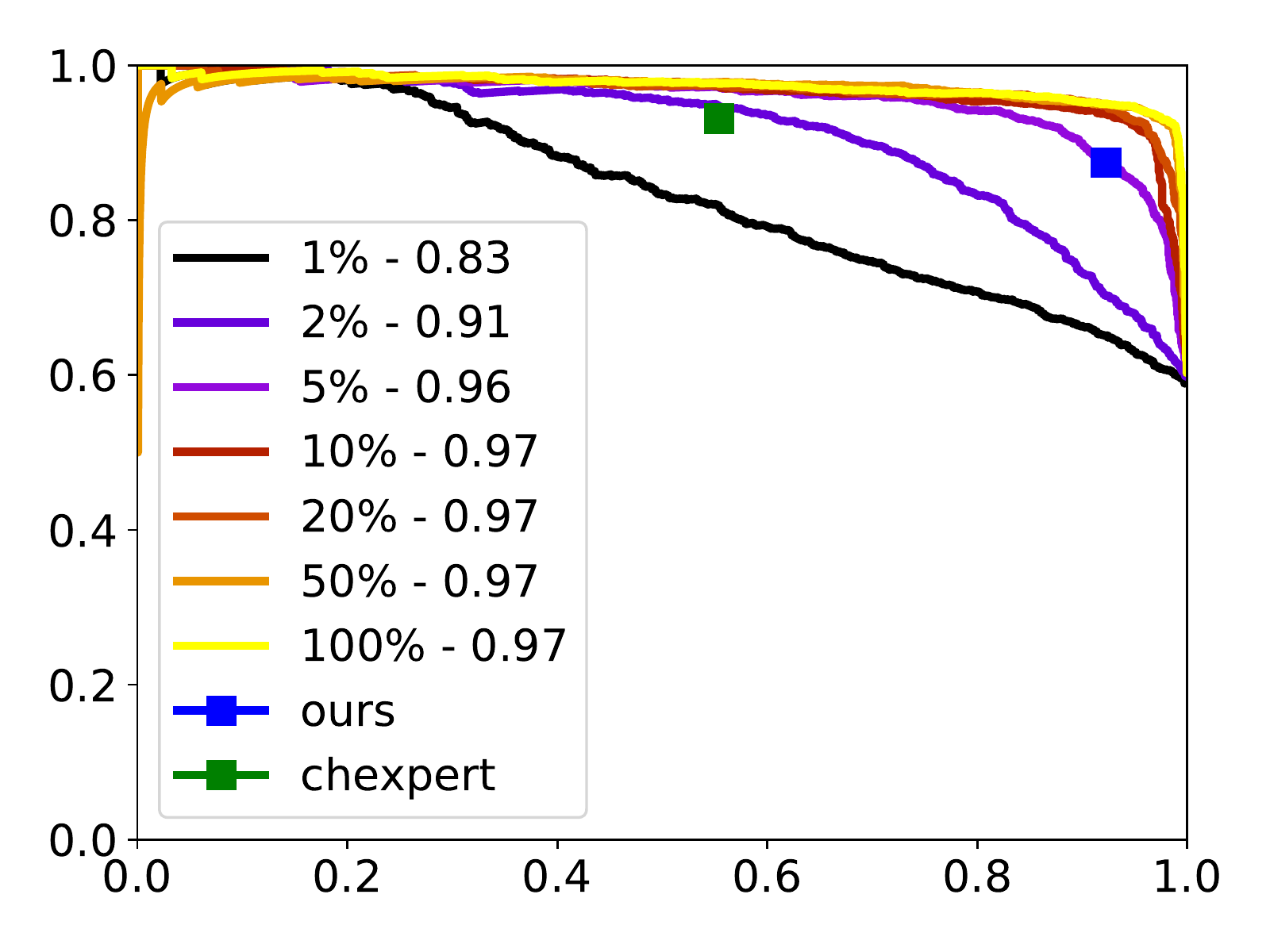}
\caption{Support Devices}
\label{sfig:rcnn_support}
\end{subfigure}\par\medskip

\caption{RCNN per-label PR curves on MIMIC test set with different data sizes and comparison with rule-based performance. As commonly shown, the y axis is Precision while the x axis is Recall. Just like the previous models, the RCNN outperforms the rule-base baseline with 20\% of the data, and obtains a better performance than other models with a small amount of labels.}
\label{fig:rcnn_per_label}
\end{figure*}

\begin{figure*}
\centering
\begin{subfigure}{.25\textwidth}
\centering
\includegraphics[width=\linewidth]{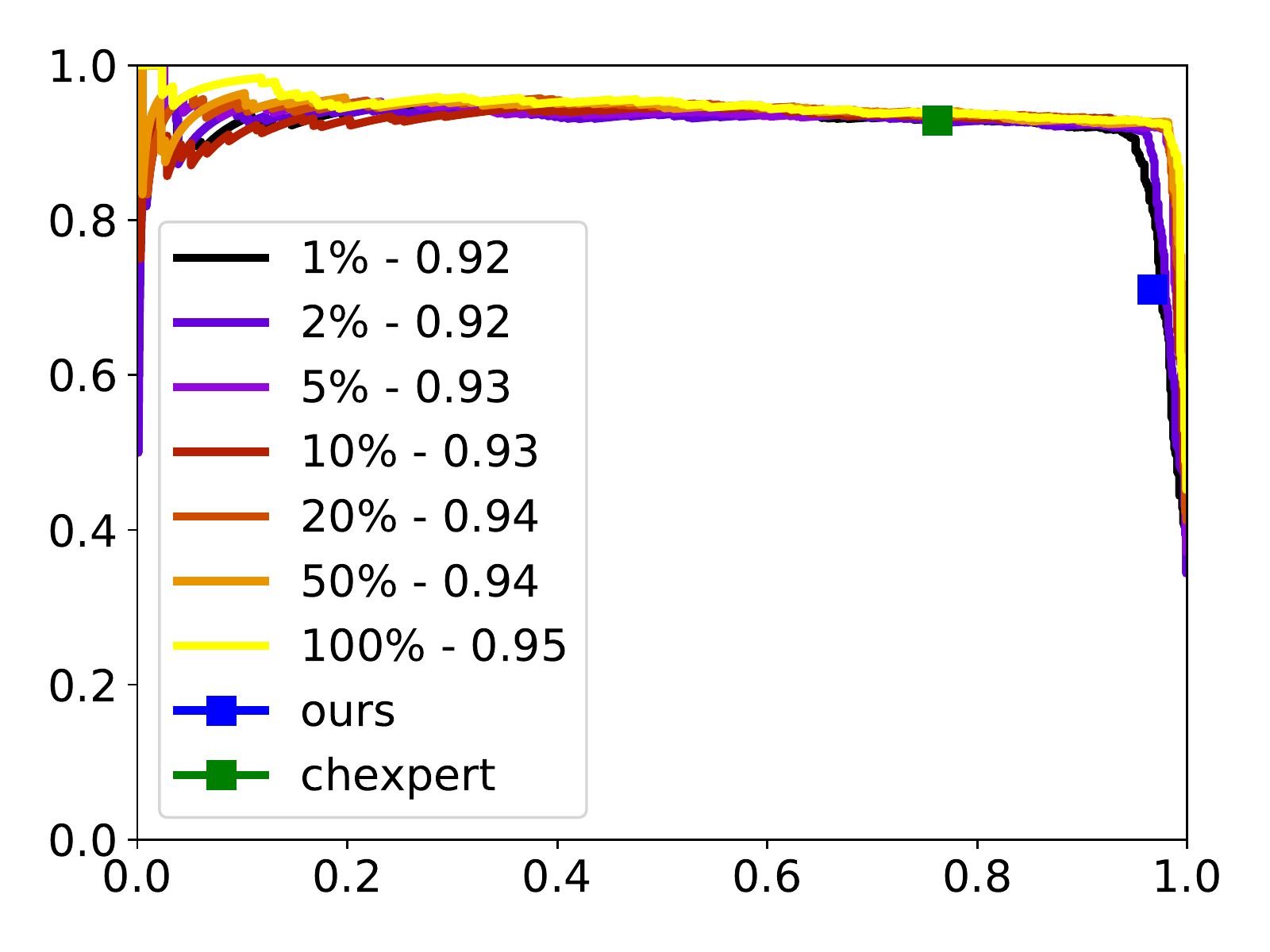}
\caption{Atelectasis}
\label{sfig:bert_atelectasis}
\end{subfigure}%
\begin{subfigure}{.25\textwidth}
\centering
\includegraphics[width=\linewidth]{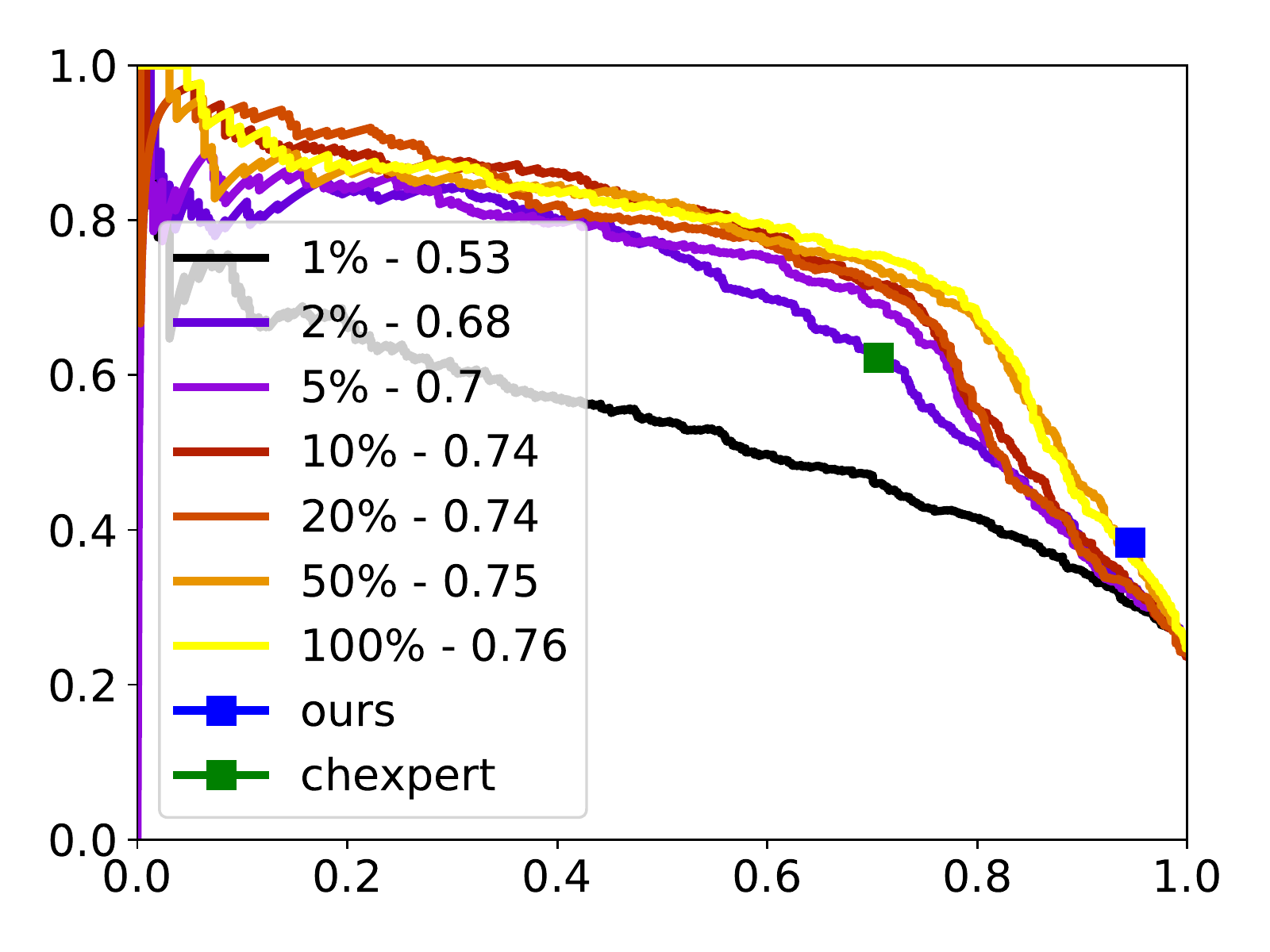}
\caption{Consolidation/Pneumonia}
\label{sfig:bert_consolidation}
\end{subfigure}%
\begin{subfigure}{.25\linewidth}
\centering
\includegraphics[width=\linewidth]{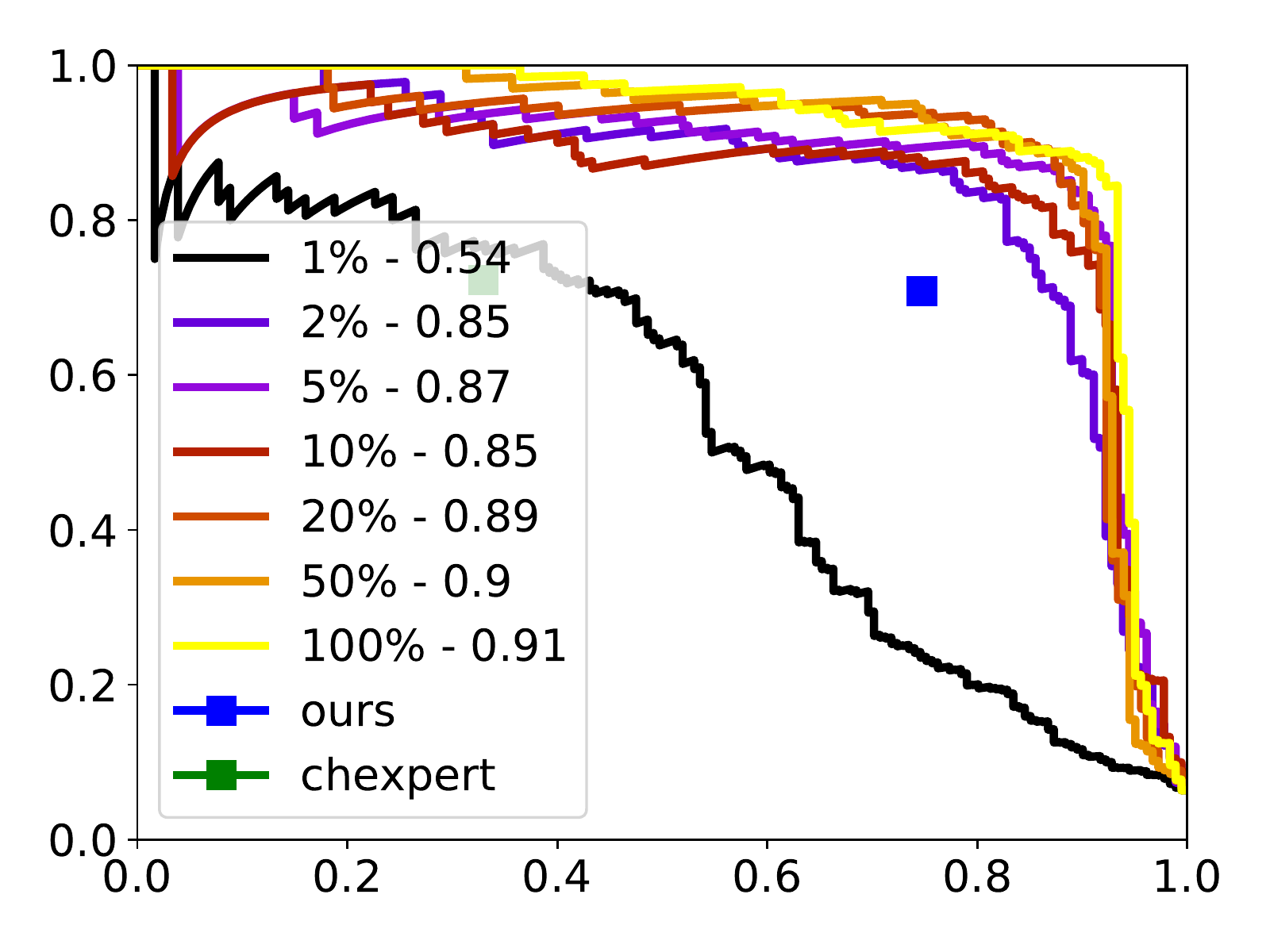}
\caption{Fracture}
\label{sfig:bert_fracture}
\end{subfigure}%
\begin{subfigure}{.25\textwidth}
\centering
\includegraphics[width=\linewidth]{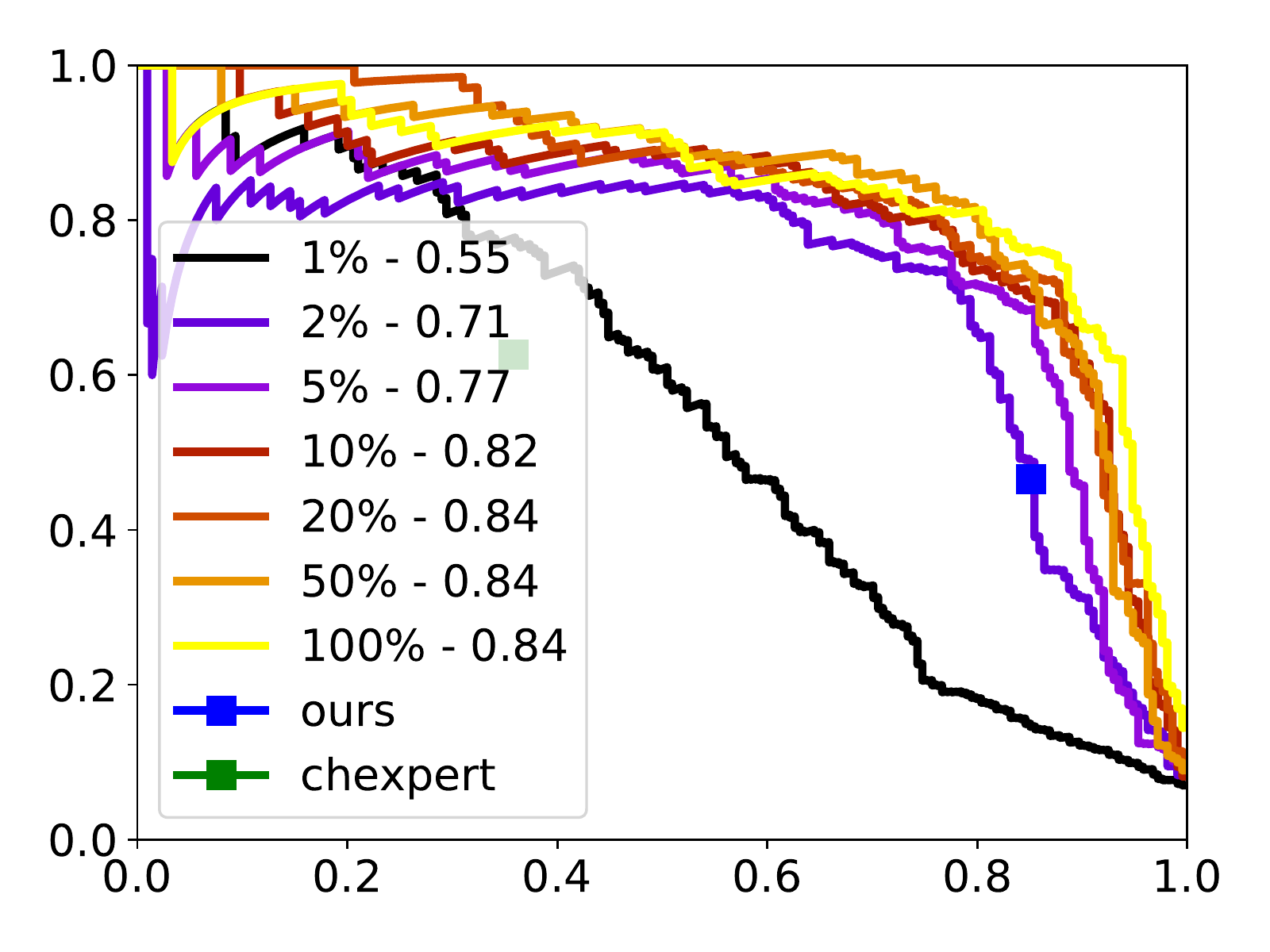}
\caption{Lung Lesion}
\label{sfig:bert_lesion}
\end{subfigure}\par\medskip
\begin{subfigure}{.25\textwidth}
\centering
\includegraphics[width=\linewidth]{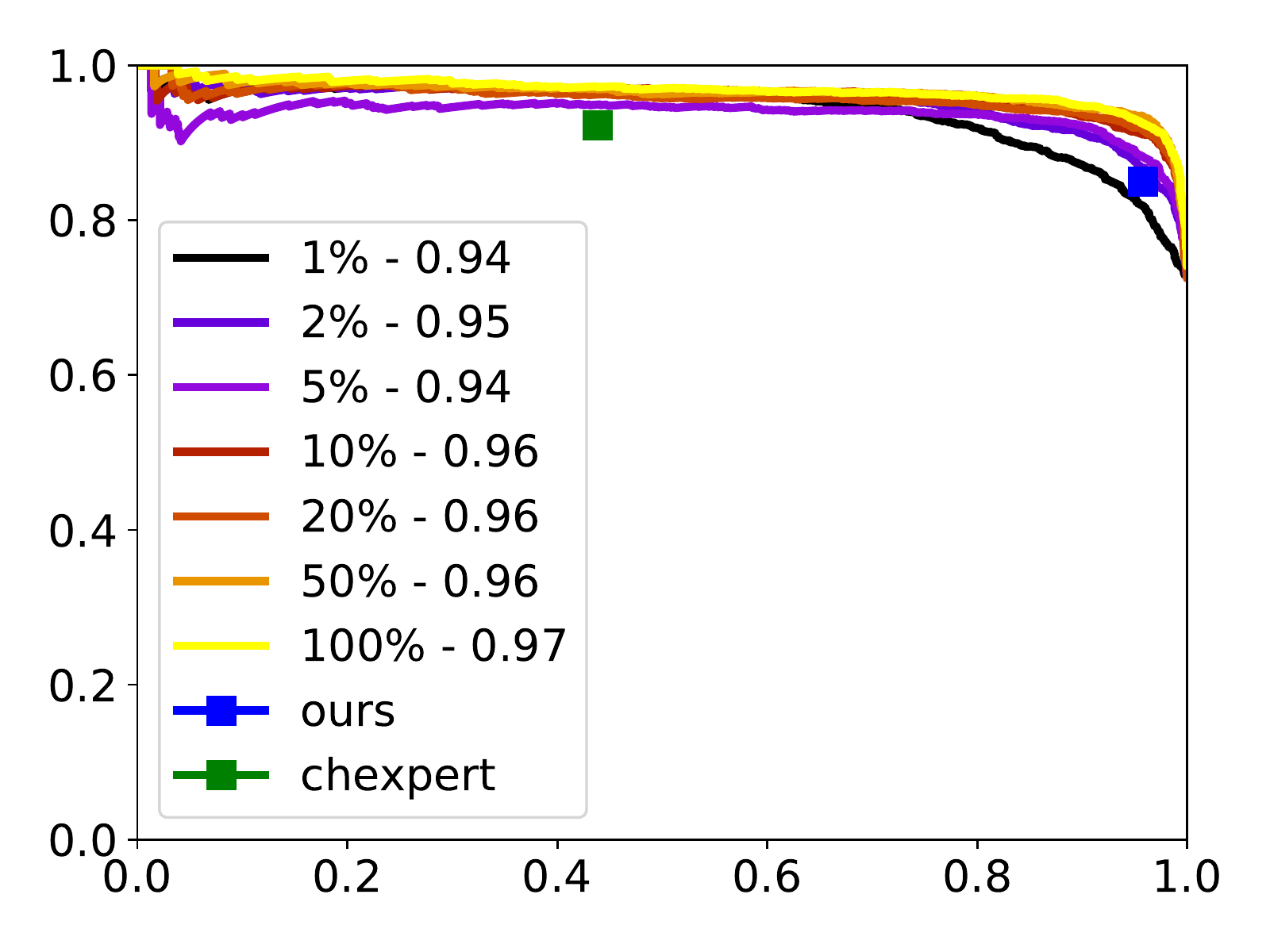}
\caption{Lung Opacity}
\label{sfig:bert_opacity}
\end{subfigure}%
\begin{subfigure}{.25\linewidth}
\centering
\includegraphics[width=\linewidth]{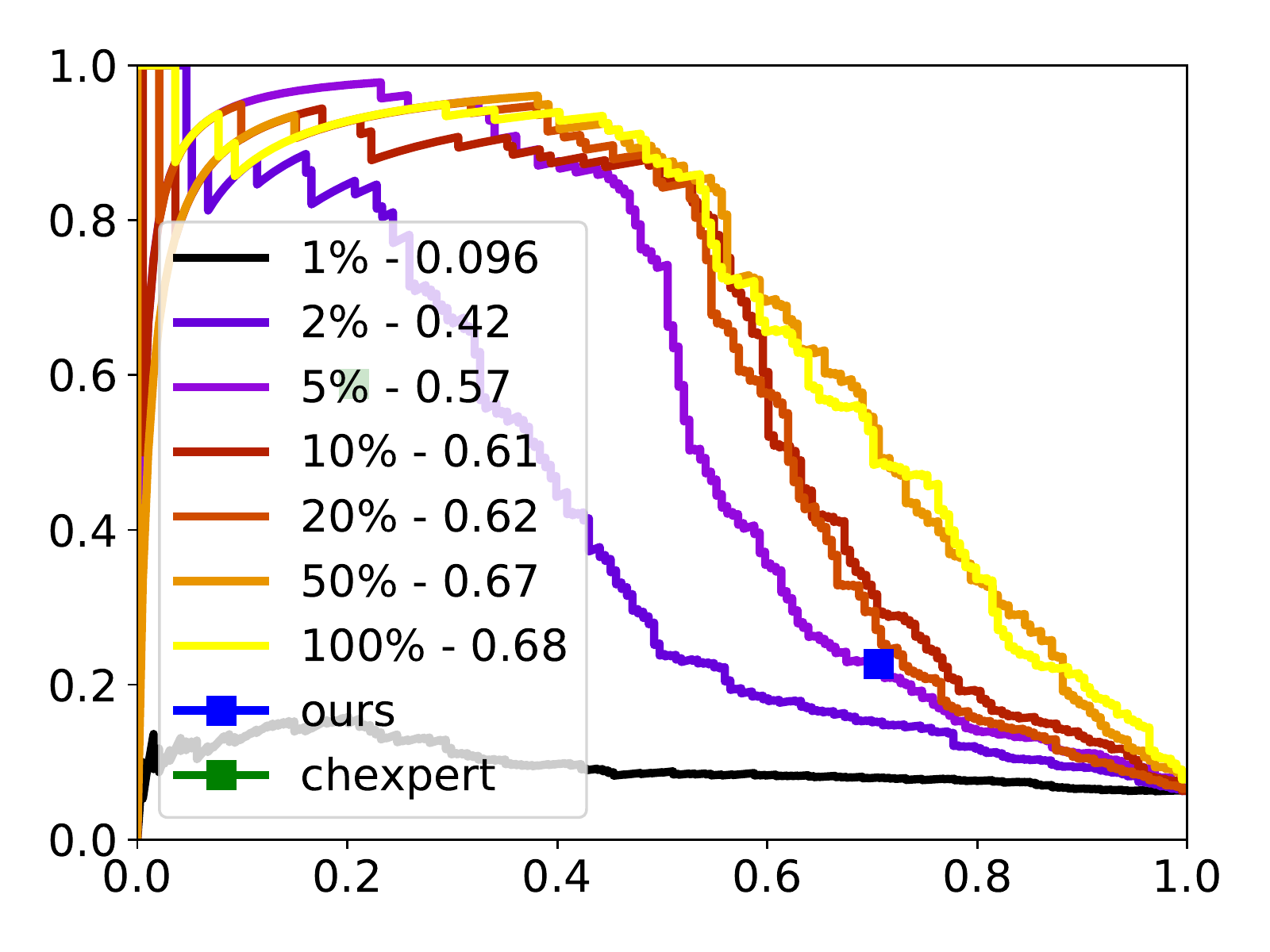}
\caption{Pleural Other}
\label{sfig:bert_pleural}
\end{subfigure}%
\begin{subfigure}{.25\linewidth}
\centering
\includegraphics[width=\linewidth]{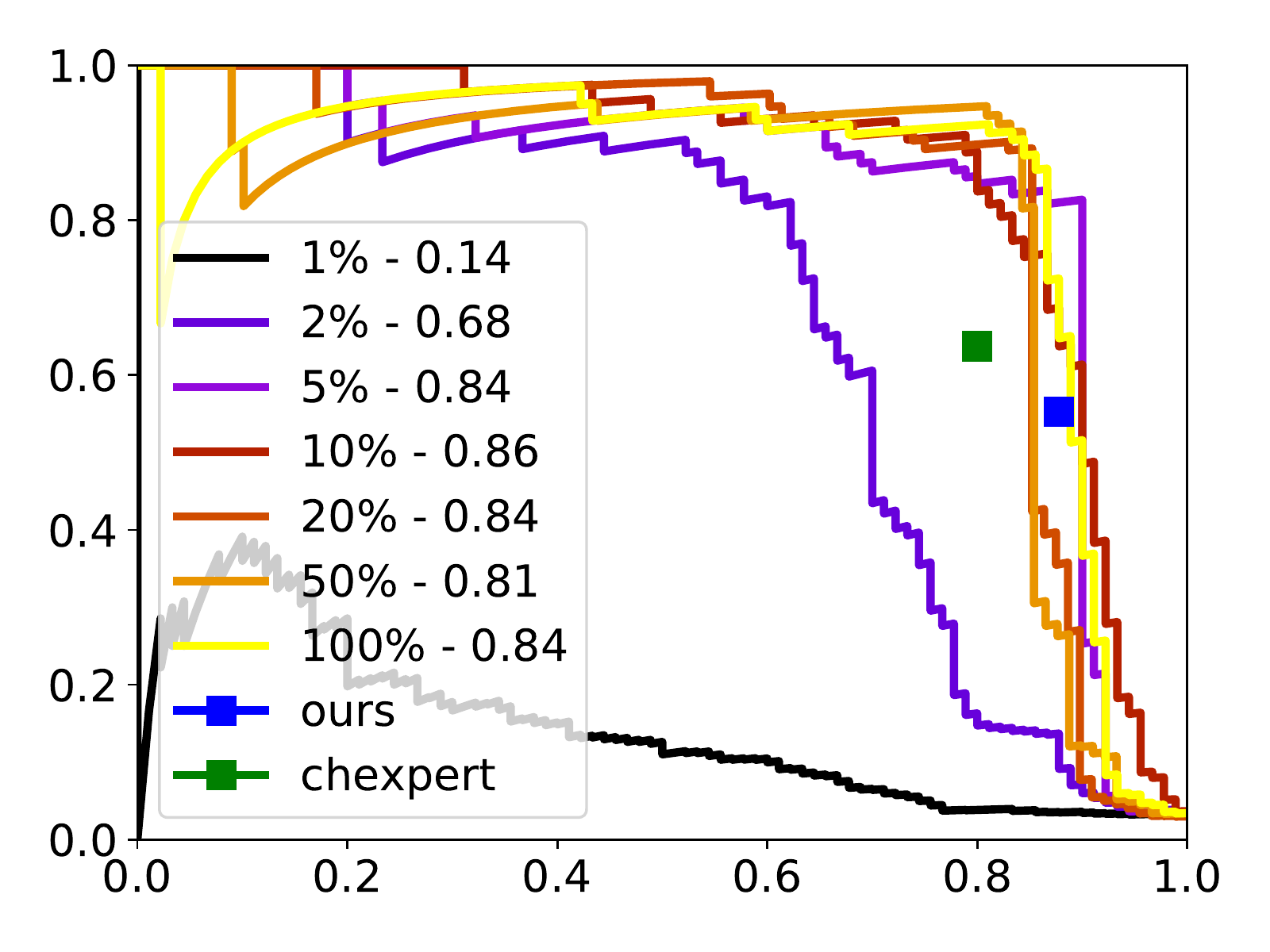}
\caption{Pneumothorax}
\label{sfig:bert_pneumothorax}
\end{subfigure}%
\begin{subfigure}{.25\textwidth}
\centering
\includegraphics[width=\linewidth]{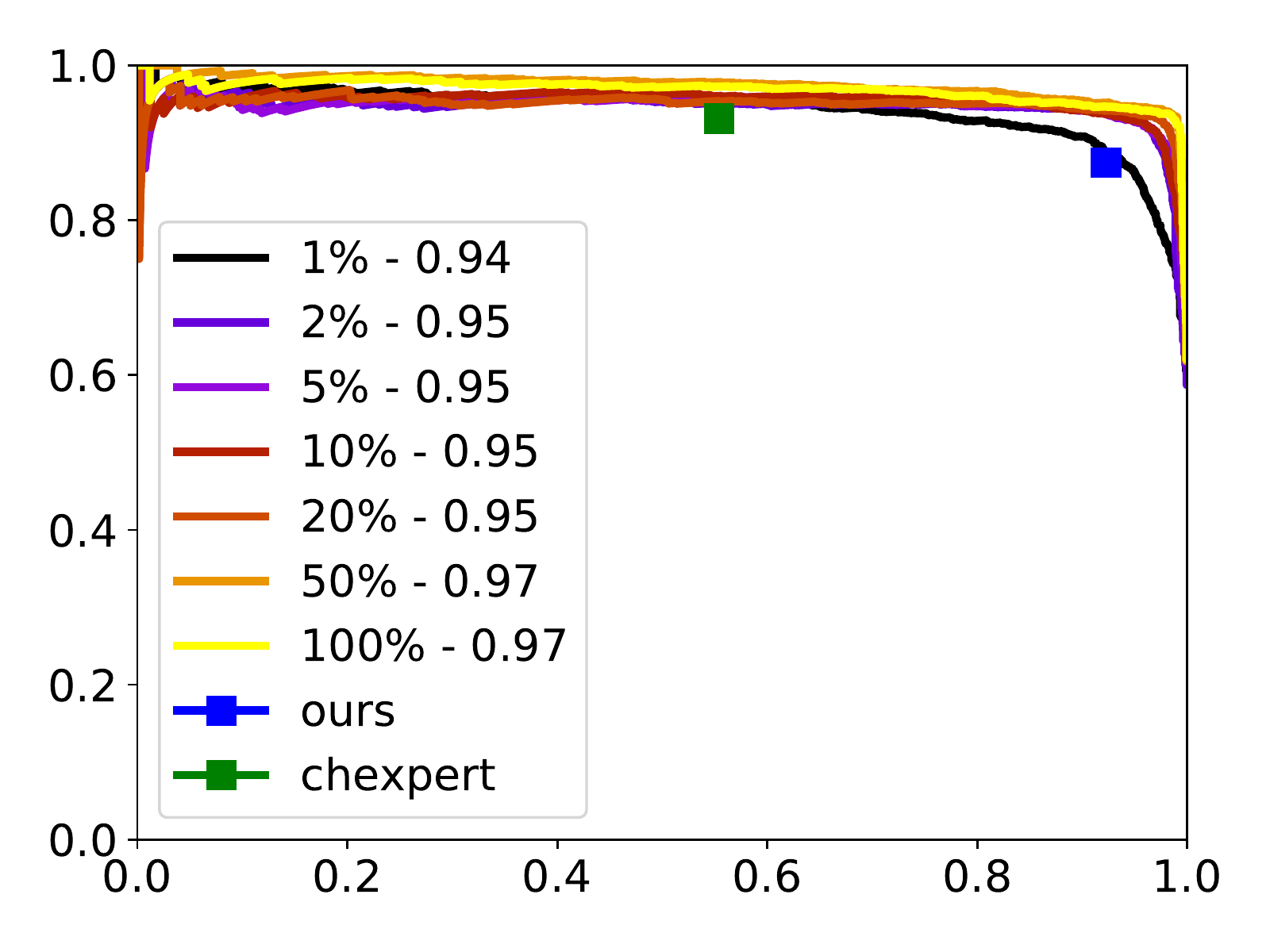}
\caption{Support Devices}
\label{sfig:bert_support}
\end{subfigure}\par\medskip

\caption{BERT per-label PR curves on MIMIC test set with different data sizes and comparison with rule-based performance. As commonly shown, the y axis is Precision while the x axis is Recall. We see here the power of the model size, architecture and pretraining. BERT learns even faster than all of the previous models and outperforms the rule-base baseline with often less than 10\% of the data.}
\label{fig:bert_per_label}
\end{figure*}

In Figures \ref{fig:rnn_per_label}, \ref{fig:cnn_per_label}, \ref{fig:rcnn_per_label} and \ref{fig:bert_per_label}, we show the Precision-Recall curves with different model architectures and dataset sizes for 8 of the 10 labels due to space limitation. The two labels that we do not show are \textit{Enlarged Cardiomediastinum} and \textit{Pleural Effusion} which are two of the best performing labels, with AUC-PRs of 0.92 and 0.97 respectively with BERT and 2\% of the data. A critical look at per category analysis as presented in the different graphs reveals a trend consistent with our intution about the problem. 
AUC-PR is poor below 5\% of the data for categories like \textit{Lung Lesion} (nodules),\textit{ Pneumothorax}, \textit{Consolidation}, \textit{Pleural Other}, and \textit{Fracture} perform poorly while being relatively higher for other categories. This contrasts with performance on categories like \textit{Support Devices}, \textit{Lung Opacity} and \textit{Atelectasis}. This trend reflects the ease with which the models are able to understand these concepts and disambiguate them.
There are a number of possible explanations for these trends with some examples in Table \ref{tab_examples} below.
\begin{itemize}
\item These hard categories are typically described syntactically and semantically with a lot of uncertainty or hedging in reports, making it difficult to determine if the concept is being affirmed or negated. The reporting radiologists is unclear as to if the abnormality in question is present or absent. The report typically requests further confirmatory investigations.
\item These hard categories reflect the inherent difficulty with visually identifying these categories on an image. For example, a \textit{Pneumothorax} is typically small and located in the crowded lung apex area. \textit{Lung lesions} exist on a spectrum from focal to diffuse or multifocal, making their descriptions less consistent in reports. 
\item Some specific categories are commonly conflated because of significant visual similarity. A known pair that reflects this phenomenon is \textit{Collapse/Consolidation}. This expression is very common in reports.
\item \textit{Pleural Other} is a catch-all category for an unspecified number of abnormalities involving the pleural cavity. There may not be sufficient examples of each of the  patterns. This pattern stays consistent even when trained with the fine-tuned BioBERT model. Vague problem definition caps the performance of even the best DL models.
\item \textit{Fracture} also represents the combination of a wide range of syntactically inconsistent abnormalities ranging from gross rib to subtle senile  or pathologic spine fractures. A more precise definition (for example disambiguating fracture types) may likely improve performance on this label.
\end{itemize}


\begin{table}[]
\small
\centering

\begin{tabular}{| m{20em} |}
\hline
Two focal areas of subsegmental \textbf{collapse/consolidation} in the left upper lobe most likely infective in nature rather than pulmonary embolus or \
inhaled foreign body. \\ \hline

There is a small calcific opacity at the right pulmonary subapical area
which is consistent with a granuloma. There is pleural thickening at the lateral aspect of the basal right hemithorax.  There is no current evidence of pleural effusion and there is no consolidation.  \\ \hline

9-mm opacity just to the right of the trachea is probably a vessel on-end, although could also be a nodule.  There may be a trace right pleural effusion on the lateral view.  No large pleural effusion or pneumothorax. \\ \hline

\end{tabular}
\caption{Report examples from the training set used in this study. They are reports dictated by radiologists and transcribed thereafter by technicians. All reports refer to chest x-rays studies.}
\label{tab_examples}
\end{table}

\begin{figure*}[ht]
\centering
\begin{subfigure}{.45\textwidth}
\centering
\includegraphics[width=\linewidth]{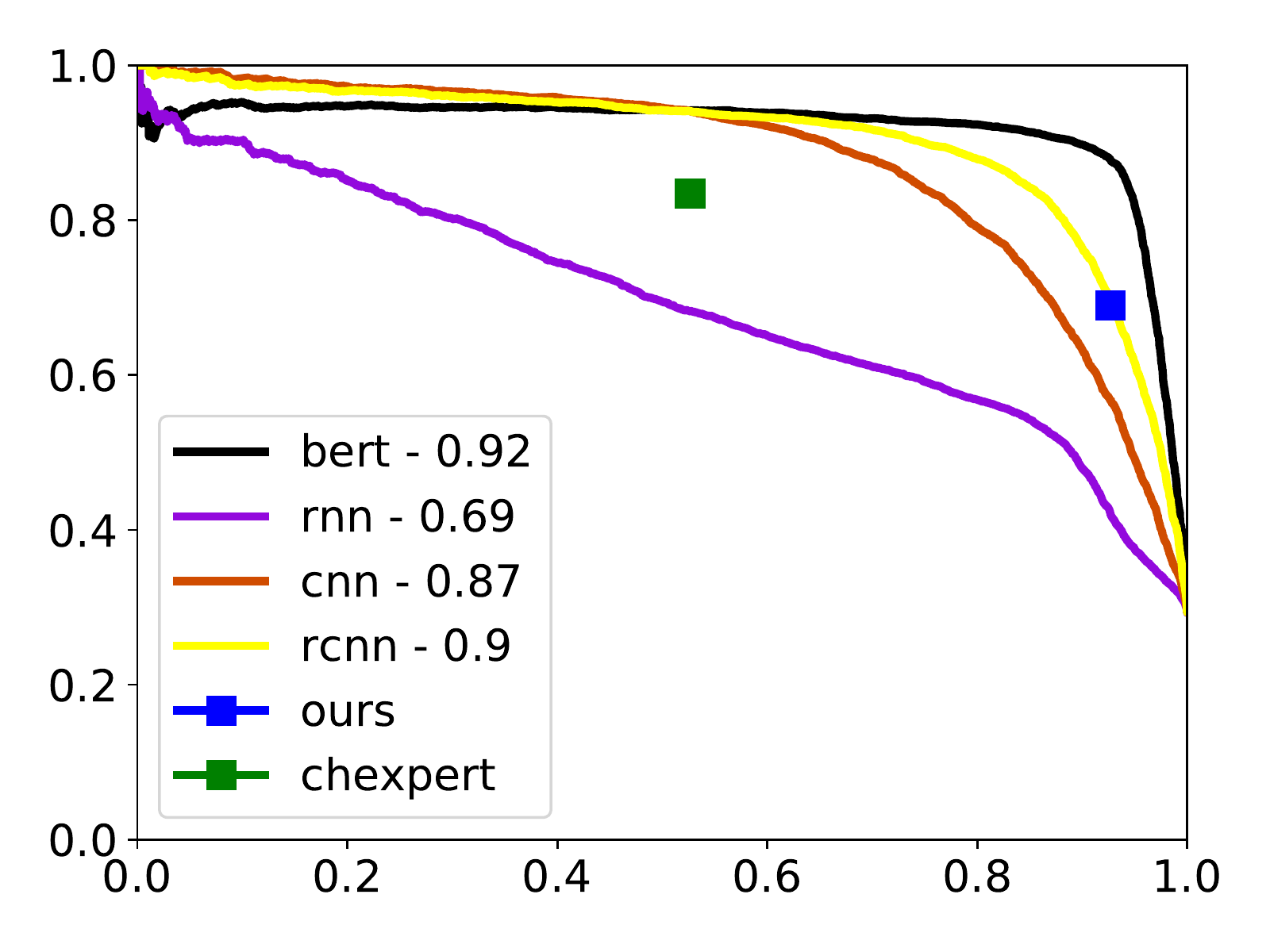}
\caption{With 5\% of the training data}
\label{sfig:5percent}
\end{subfigure}%
\begin{subfigure}{.45\textwidth}
\centering
\includegraphics[width=\linewidth]{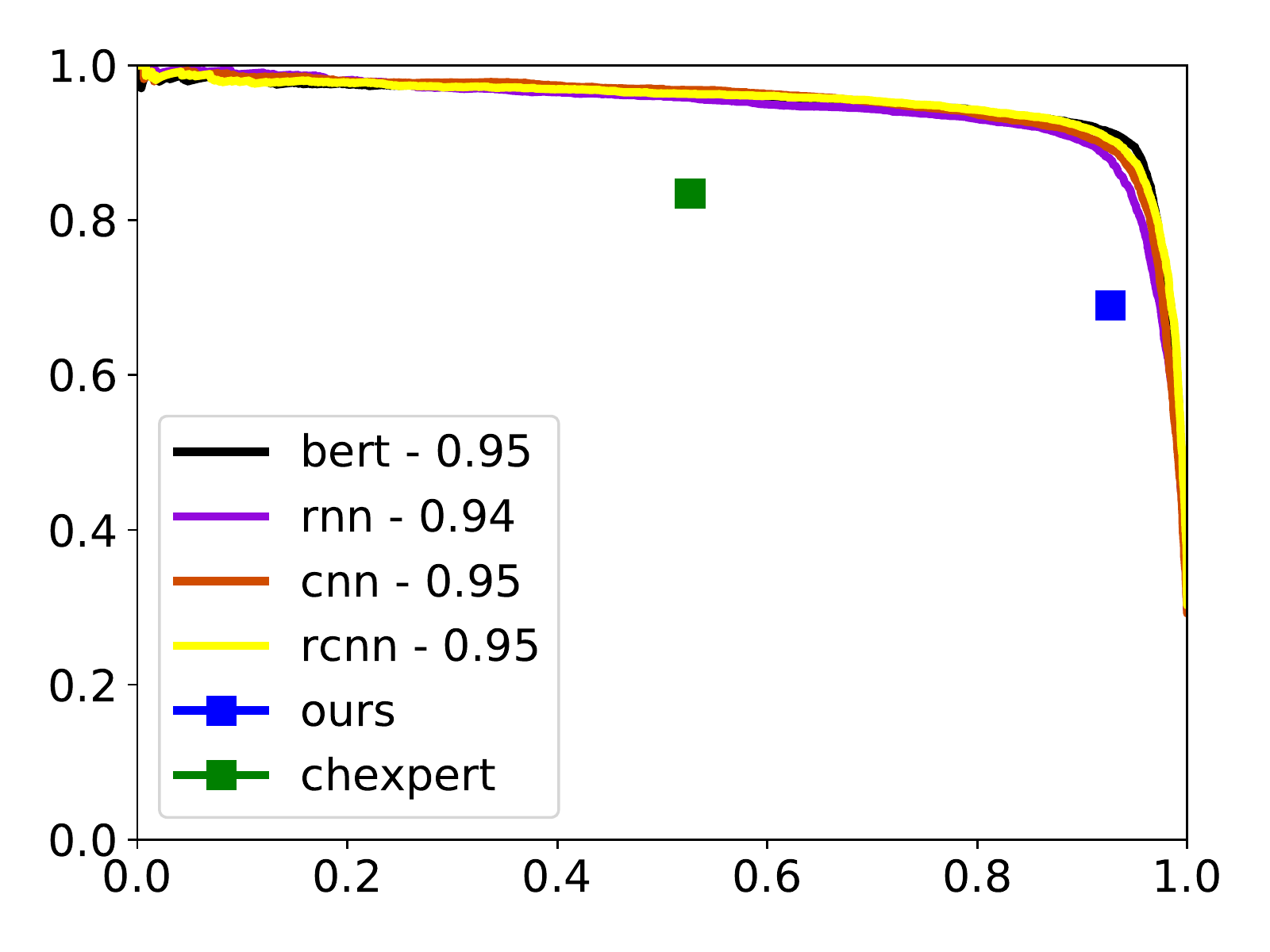}
\caption{With 100\% of the training data}
\label{sfig:100percent}
\end{subfigure}
\caption{PR curves on the MIMIC test set for each model and comparison with rule-based performance}
\label{fig:models}
\end{figure*}

\subsection{Effect of model architecture and pretraining}

Figure \ref{fig:models} shows the comparison of the model performance across all four architectures on 5\% of the data and the full dataset. Across the selected model architectures, performance is comparable at 100\% of the data. As training dataset shrinks, important distinctions begin to emerge. The intuition here is that the maxpooling in the CNNs (CNN, RCNN) serve as efficient keyword detectors even with small data, whereas the LSTM, despite the attention mechanism, struggles to match this performance because it requires a lot more data to learn how to identify important keywords from the entire sequence in the forward and backward directions. The RCNN outperforms the CNN because of the additional recurrence on top of pooling capabilities. BioBERT however holds steady over 0.90 even at 2\% of the data. This pattern reflects the extensive clinical, but also English knowledge contained in the BioBERT model carried over from pretraining tasks on large clinical corpus. The knowledge proves useful even at 1\% of the data where the RNN, trained from scratch, does slightly better than a coin toss on MIMIC, and far worse on OpenI.

\subsection{Effect of the ratio of positive reports}
\begin{figure*}[ht]
\centering
\includegraphics[width=.7\linewidth]{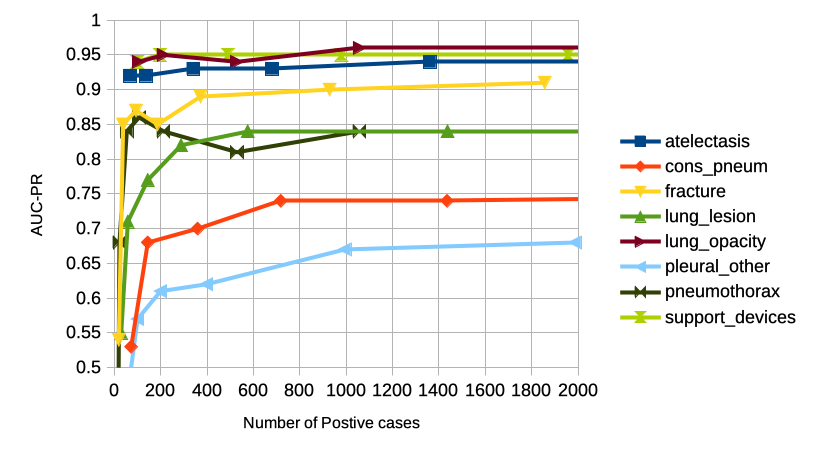}
\caption{AUC-PR curves of the BioBERT model with increasing number of positive cases for each label. The right side of the curves are cropped in order to focus on the area of change.}
\label{fig:pos_cases}
\end{figure*}
The number of reports itself is not the unique influence on the performance of the model. As we discussed previously, it also heavily depends on the labels and their definition, as well as the model architecture and its pretraining. We tackle in this paragraph another important element, the ratio of positive reports for each label. Labeling 6,000 reports will not help the model if all of them are negative for the label we are interested in. Figure \ref{fig:pos_cases} shows a different take on the curves that we have previously shown. Instead of highlighting the impact of the number of reports on the performance of the model, we plot instead the impact of the number of positive reports for each label. This is particularly important because each label having its own definition means that: i) the labels are not equally likely to be present in a random subset of the data and ii) they may not need the same amount of training data, depending on their difficulty. Figure \ref{fig:pos_cases} shows once again that some labels are more difficult than others. Additionally, some labels require a smaller amount of data for the performance to converge. However, these curves allow us to quantitatively conclude that for this task, 600 positive reports per label seems to lead to an early performance saturation regardless of abnormality types. Adding more positive cases beyond this boundary leads to diminishing returns.

\subsection{Generalization to other tasks}
Finally, we want to mention some limitations to our conclusions. While we believe that these findings hold true for any kind of classification on radiology reports (not just chest X-ray), radiology reports only represent a subset of the clinical text data. They are written with a clear objective in mind, sometimes using templates and thus resulting in a smaller vocabulary than random free text. Additionally, if the task was Named Entity Recognition (NER) instead of classification for example, then the results might be slightly different. However, we believe that the main reason behind our findings lies in the uniqueness of clinical data. Clinical data is based on well defined and documented knowledge and experience, therefore containing highly templated syntactic and semantic patterns compared with casual conversations. Due to this nature, we think that our conclusions would still be valid to a large extent in a different clinical NLP context provided that the choice of pretraining and finetuning methods is appropriate.

\section{Conclusion}
The intuition from low resource languages and the general non-medical domain suggests sustained performance improvement with larger datasets. We seek to answer the lingering question about how much data is considered as sufficient in classifying clinical texts and show that the necessary corpus size varies with the complexity of the specific category but to a far less degree than what has been previously thought. With less than 6,000 labeled reports, DL models are able to yield comparable performance with using 30,000 reports on a multilabel classification problem, demonstrating the counter-intuitive effect of diminishing returns from expert labeling efforts.

\bibliography{emnlp2020}
\bibliographystyle{acl_natbib}
\end{document}